\documentclass[aoas]{imsart}
\RequirePackage{amsthm,amsmath,amsfonts,amssymb}
\RequirePackage[authoryear]{natbib}
\usepackage[dvipsnames]{xcolor}
\usepackage{enumerate}
\usepackage{graphicx}
\usepackage{amsmath}
\usepackage{booktabs}
\usepackage[flushleft]{threeparttable}
\usepackage{bbold}
\usepackage{amsthm}
\usepackage{lineno}
\usepackage[utf8]{inputenc}

\usepackage{array}
\usepackage{makecell}
\usepackage[ruled]{algorithm2e}
\usepackage{amssymb}
\usepackage{mdframed}
\usepackage{cellspace}
\usepackage{soul}
\usepackage{framed}
\usepackage[title]{appendix}
\usepackage[normalem]{ulem}
\usepackage{bbm}
\usepackage[english]{babel}
\RequirePackage[colorlinks,citecolor=blue,urlcolor=blue]{hyperref}
\usepackage{float}
\usepackage[caption=false,listofformat=subsimple,labelformat=simple]{subfig}

\usepackage{multirow}

\startlocaldefs
\theoremstyle{plain}

\theoremstyle{remark}
\newtheorem{example}{Example}
\newcolumntype{?}{!{\vrule width 0.8pt}}

\usepackage{centernot}

\newcommand*{\threeemdash}{\rule[0.5ex]{3em}{0.8pt}}

\newtheorem{innercustomgeneric}{\customgenericname}
\providecommand{\customgenericname}{}
\newcommand{\newcustomtheorem}[2]{%
  \newenvironment{#1}[1]
  {%
   \renewcommand\customgenericname{#2}%
   \renewcommand\theinnercustomgeneric{##1}%
   \innercustomgeneric
  }
  {\endinnercustomgeneric}
}

\makeatletter
\renewcommand\paragraph{\@startsection{paragraph}{4}{\z@}%
                                     {-3.25ex\@plus -1ex \@minus -.2ex}%
                                     {1.5ex \@plus .2ex}%
                                     {\normalfont\normalsize\bfseries}}
\makeatother

\makeatletter
\newcommand*\bigcdot{\mathpalette\bigcdot@{.5}}
\newcommand*\bigcdot@[2]{\mathbin{\vcenter{\hbox{\scalebox{#2}{$\m@th#1\bullet$}}}}}
\makeatother

\newcustomtheorem{customthm}{Theorem}
\newcustomtheorem{customlemma}{Lemma}

\newcommand{\argmin}{\mathop{\mathrm{argmin}}}

\endlocaldefs

\begin{document}

\begin{frontmatter}
\title{Nonparametric Additive Value Functions: \\ Interpretable Reinforcement Learning with an \\ Application to Surgical Recovery}
\runtitle{Nonparametric Additive Value Functions}

\begin{aug}
\author[A]{\fnms{Patrick}~\snm{Emedom-Nnamdi}\ead[label=e1]{patrickemedom@hsph.harvard.edu}},
\author[B]{\fnms{Timothy}~\snm{R. Smith}},
\author[A]{\fnms{Jukka-Pekka}~\snm{Onnela}}\and\author[A]{\fnms{Junwei}~\snm{Lu}}
\address[A]{Department of Biostatistics, Harvard University,\\\printead[presep={\ }]{e1}}

\address[B]{Computational Neurosciences Outcomes Center, Department of Neurosurgery, \\Brigham and Women's Hospital, Harvard Medical School\printead[presep={\ }]{}}
\end{aug}

\begin{abstract}

We propose a nonparametric additive model for estimating interpretable value functions in reinforcement learning, with an application in optimizing postoperative recovery through personalized, adaptive recommendations. While reinforcement learning has achieved significant success in various domains, recent methods often rely on black-box approaches such as neural networks, which hinder the examination of individual feature contributions to a decision-making policy. Our novel method offers a flexible technique for estimating action-value functions without explicit parametric assumptions, overcoming the limitations of the linearity assumption of classical algorithms. By incorporating local kernel regression and basis expansion, we obtain a sparse, additive representation of the action-value function, enabling local approximation and retrieval of nonlinear, independent contributions of select state features and the interactions between joint feature pairs. We validate our approach through a simulation study and apply it to spine disease recovery, uncovering recommendations aligned with clinical knowledge. This method bridges the gap between flexible machine learning techniques and the interpretability required in healthcare applications, paving the way for more personalized interventions.
\end{abstract}


\end{frontmatter}

\section{Introduction}

Optimizing postoperative recovery is crucial for improving surgical outcomes, including functional restoration and enhanced quality of life. This process is inherently complex and multifaceted, influenced by the nature of the diagnosis, the type of surgical procedure, and various patient-specific factors. Demographic characteristics such as age, gender, and comorbidities, as well as time since surgery and patient behaviors (e.g., mobility, physical activity, and sleep) significantly impact recovery trajectories \citep{Cote2019, Panda2020SmartphoneSurgery}. While common recommendations often focus on early mobilization activities, such as getting out of bed or taking light walks, developing personalized care plans presents a greater challenge. To provide tailored suggestions that account for diverse patient factors and recommend specific actions (e.g., daily step targets), clinicians require both continuous access to patient health and behavioral data and flexible algorithms capable of processing this information and providing interpretable insights.

The proliferation of smartphones and wearable devices has revolutionized our ability to collect real-time, continuous data on human behavior and health, providing valuable insights into patient recovery \citep{Torous2015a, Onnela2020OpportunitiesData}. This mobile health approach, when combined with statistical machine learning techniques, offers clinicians a powerful paradigm for uncovering nuanced patterns in recovery trajectories and developing more refined, evidence-based standards of care. By leveraging high-quality, temporally-dense data, healthcare professionals can now enhance their understanding of postoperative recovery and pave the way for more personalized, adaptive, and effective treatment strategies \citep{Panda2020UsingSurgery}.
 \begin{figure}[t]
   \centering
   \includegraphics[width=0.75\linewidth]{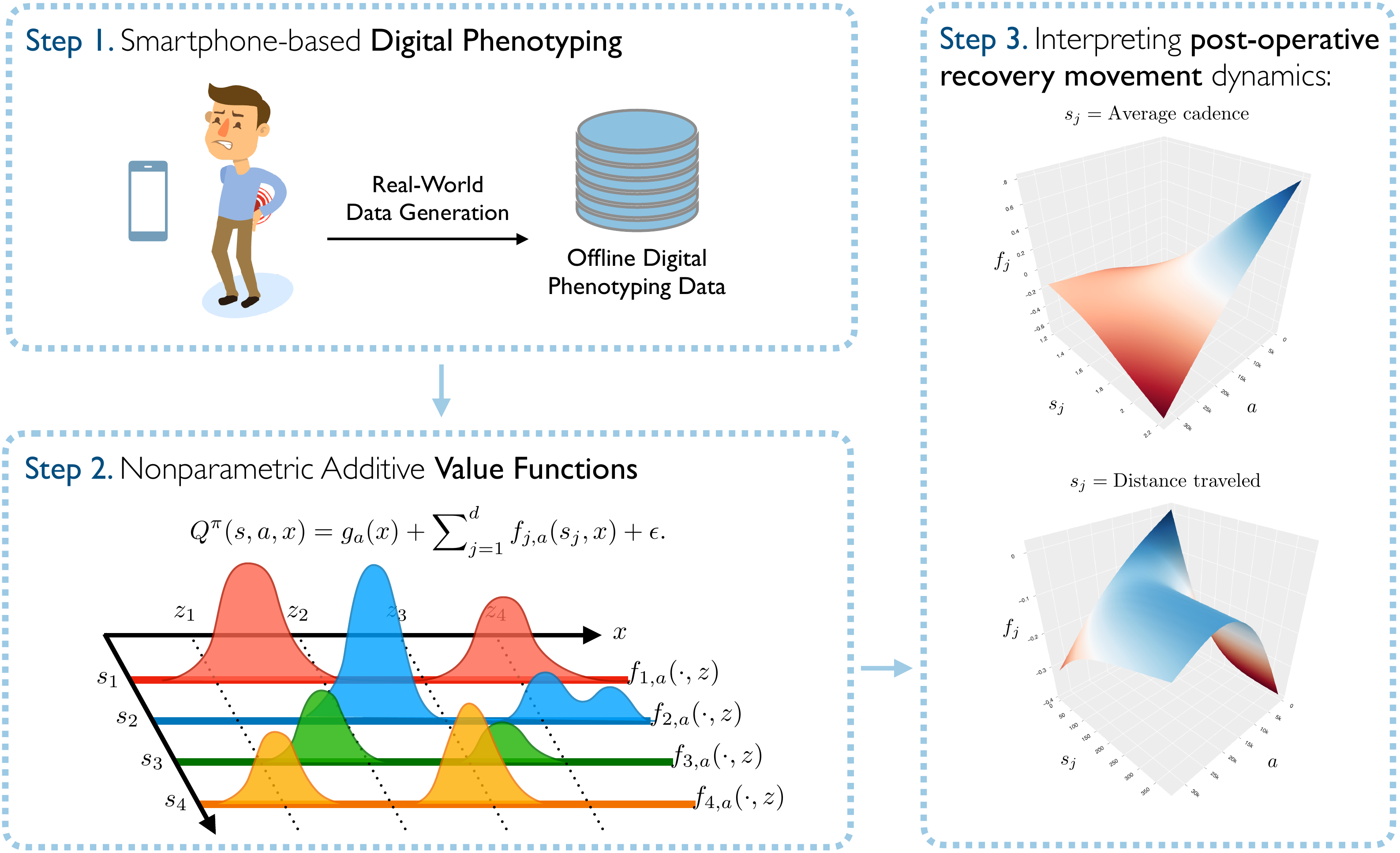} \\
   \caption{\footnotesize\textit{An overview of using nonparametric additive models for learning interpretable value functions. Within our setting, real-world data from subjects with select physiological disorders are collected using a smartphone-based mobile health platform. Modalities collected from subject smartphones range from raw sensor data (e.g., GPS, accelerometer, gyroscope, or magnetometer) to usage logs (e.g., anonymized communication and screen time). Relevant features $\mathbf{s} = (s_1,\dots,s_d)^T$ are summarized from these modalities and are used to frame a corresponding decision-making problem (or MDP see \ref{Section 2.1}) of interest. Under the select MDP, we model the value function $Q^\pi(\mathbf{s},a, x)$ as a sum of nonparametric component functions $g_a(x)$ and $f_{j,a}(\mathbf{s}_j,x)\hspace{3pt}\forall j$. Here we visualize the change in the shape and sparsity patterns of the component function $f_{j,a}(\cdot,z)$ as the candidate variable $x$ changes, e.g., $x\in\{z_1, z_2, z_3, z_4\}$. Each additive component function can be estimated and inspected using the kernel-weighted least square fixed point approximation detailed in \ref{KSH-LSPI Section}.}} 
 \end{figure} 
 
In this paper, we introduce a novel reinforcement learning approach for estimating recovery strategies and recommendations using mobile health data. Reinforcement learning is a sub-field of machine learning that focuses on learning sequences of decisions that optimize long-term outcomes from experiential data \citep{Sutton2018ReinforcementDraft}. In healthcare, reinforcement learning algorithms have been used to discover decision-making strategies for chronic disease treatments \citep{Bothe2013ThePancreas, Peyser2014TheDiabetes}, anesthesia regulation and automation \citep{Sinzinger2011SedationControl}, chemotherapy scheduling and dosage management \citep{Padmanabhan2015Closed-loopLearning, Ahn2011DrugApproach}, and sepsis management \citep{Raghu2017ContinuousApproach, Peng2018ImprovingLearning}. 

Implementing reinforcement learning in healthcare applications requires careful consideration of the policy estimation process and thorough examination of the learned policy's behavior prior to real-world deployment \citep{Gottesman2019GuidelinesHealthcare}. Typically, decision-making policies are represented as a function $\pi(\mathbf{s})$ of state features $\mathbf{s} = (s_1,\dots,s_d)^T \in \mathbb{R}^d$, estimated using policy gradient or value-based reinforcement learning algorithms \citep{Sutton2018ReinforcementDraft}. In value-based approaches, policies are determined by selecting the action $a$ that maximizes the corresponding action-value function $Q^\pi(\mathbf{s},a)$. Current methods often employ neural network function approximators, resulting in black-box algorithms that are difficult to interpret and provide minimal insight into which feature (or set of features) influenced the decision-making process \citep{Gottesman2019GuidelinesHealthcare}. While linear models offer interpretability, they fail to capture the complex, nonlinear relationships and interactions between clinical variables that are crucial for understanding and optimizing surgical recovery, necessitating a more flexible and interpretable approach.

To address these limitations, we propose a novel class of approximate value functions that offers a flexible, nonparametric representation of the action-value function, easily interpretable for both researchers and clinicians. Our approach allows for the inspection of a candidate variable $x$ (e.g., time-varying/-invariant confounders or continuous-valued actions) and models action-value functions as a sum of nonparametric, additive component functions:
\begin{align}
Q^\pi(\mathbf{s},a,x) &=  g_a(x) + \sum_{j=1}^df_{j,a}(\mathbf{s}_j,x) + \epsilon.
\label{primary-model}
\end{align}
This framework enables the examination of both the marginal effect of $x$ and its joint effect with state features $\mathbf{s}_j$ under a discretized action space. To estimate these component functions, we build upon the classical Least Square Policy Iteration (LSPI) algorithm \citep{Lagoudakis2004Least-squaresIteration}, incorporating a kernel-hybrid approach that relaxes traditional linearity assumptions. By leveraging advances in high-dimensional, nonparametric additive regression models \citep{Fan2005NonparametricModels, Ravikumar2007SparseModels, Lafferty2008Rodeo:Regression}, we introduce a kernel-sieve hybrid regression estimator \citep{Lu2020KernelModel} to obtain a sparse additive representation of the action-value function.

To validate our methodology, we present a simulation study examining its ability to estimate nonlinear additive functions and compare its performance against modern neural network-based approaches. Furthermore, we apply our model to an ongoing mobile health study, demonstrating its capacity to learn and interpret a decision-making policy aimed at improving pain management and functional recovery in patients recovering from spine surgery through mobility management.

\subsection{Related Research}

Our work contributes to the growing literature on function approximation methods for value-based reinforcement learning. Current state-of-the-art algorithms approximate action-value functions using expressive modeling architectures, such as neural networks. By combining fitted Q-iteration procedures with modern tools such as replay buffers and target networks, these algorithms resolve the pitfalls of classical methods and solve complex, high-dimensional decision-making tasks \citep{Riedmiller2005NeuralMethod, Antos2007FittedMDPs, VanHasselt2010DoubleQ-learning, Mnih2013PlayingLearning, Mnih2015Human-levelLearning}. However, the powerful flexibility of these approaches comes at the cost of interpretability inherent in algorithms such as Least Squares Policy Iteration (LSPI).

LSPI is a model free, off-policy approximate policy iteration algorithm that models the action-value function using a parametric linear approximation and finds an approximate function that best satisfies the Bellman equation (i.e., the fixed-point solution) \citep{Lagoudakis2004Least-squaresIteration}. While LSPI provides an unbiased estimate of the action-value function, it faces significant challenges when the model is misspecified and when the dimensionality of the feature space is high \citep{Lagoudakis2004Least-squaresIteration, Farahmand2016RegularizedSpaces}. Several modifications to LSPI have been proposed in the reinforcement learning literature. In settings where the feature space is large, several approaches exist for finding sparse solutions in a linear model \citep{Hoffman2012RegularizedPenalization, Kolter2009RegularizationLearning, Tziortziotis2017BayesianRegularization, Geist2011L1-penalizedResidual}. Alternatively, \citet{Xu2007Kernel-basedLearning} propose a kernel-based LSPI algorithm that operates in an infinite-dimensional Hilbert space and allows for nonlinear feature extraction by selecting appropriate kernel functions. Additionally, \citet{Howard2013LocallyEnvironments} propose a locally-weighted LSPI model that leverages locally-weighted to construct a nonlinear, global control policy. 

To date, no approximation methods in RL have been introduced that directly allows for the nonparametric estimation concerning the additive contribution of select features and joint feature pairs. In the supervised learning literature, several approaches exist for estimating nonparametric component functions in high-dimensional feature spaces. Several of these approaches include generalized additive models (GAM) and sparse additive models (SpAM), which our approach draws parallels to \citep{Ravikumar2007SparseModels, Hastie2017GeneralizedModels}. To bridge these areas of research, we re-formulate the policy evaluation step of the classical LSPI algorithm and propose incorporating the kernel-sieve hybrid regression estimator introduced in \citet{Lu2020KernelModel}. This approach provides a powerful function approximation technique for locally estimating action-value functions using a loss function combining both basis expansion and kernel method with a hybrid $\ell_1/\ell_2$-group Lasso penalty. 

\subsection{Organization of the Paper}
The remainder of this paper is organized as follows. In Section \ref{Kernel-Sieve Hybrid}, we introduce our generalized, nonparametric model for representing action-value functions. Section \ref{KSH-LSPI Section} presents our estimation strategy for locally approximating the action-value function by combining basis expansion and kernel methods. Section \ref{simulation-study} examines the results of a simulation study, highlighting the performance of our method in estimating sparse additive components of the action-value function. Section \ref{Motivating Case Study} introduces a real-world cohort of patients recovering from a neurological intervention for spine disease as a motivating case study. In Section \ref{application}, we apply our method to this mobile health study and interpret the estimated recovery strategy. Finally, Section \ref{Discussion} discusses the limitations of our method and proposes future extensions to address them.

\begin{figure}
    \centering
    \includegraphics[scale=0.34]{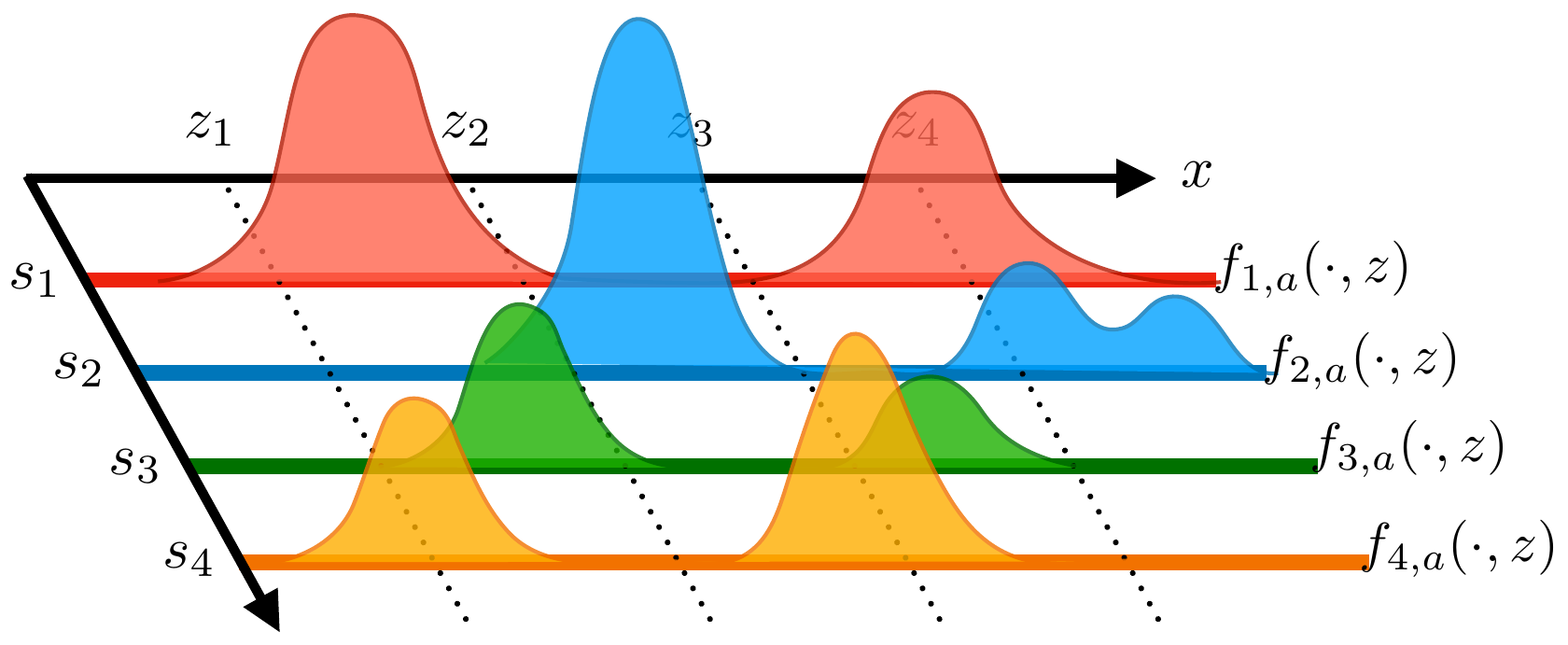}
    \caption{\footnotesize \textit{Representation of a nonparametric additive value function $Q^\pi(\mathbf{s}, a, x)$ with respect to the candidate variable $x$ as detailed in \eqref{general-model}.}}
    \label{fig:my_label}
\end{figure}

\section{Nonparametric Additive Value Functions}

\subsection{Preliminaries and Notation}
\label{Section 2.1}
We consider a discrete-time, infinite horizon Markov Decision Process (MDP) defined by the tuple $\{\mathcal{S}, \mathcal{A}, \mathcal{P}, R, \gamma\}$, where $\mathcal{S}$ is a continuous $d$-dimensional state space, $\mathcal{A}$ is a set of discrete (i.e., $\mathcal{A} = \{1,\dots,k \}$) or continuous (i.e., $\mathcal{A}=\mathbb{R}$) actions, $\mathcal{P}(\mathbf{s}'|\mathbf{s},a)$ is a next-state transition probability kernel that specifies the probability of transitioning from state $\mathbf{s} \in \mathcal{S}$ to the next state $\mathbf{s}' \in \mathcal{S}$ after taking action $a\in \mathcal{A}$, $R : \mathcal{S} \times \mathcal{A} \rightarrow \mathbb{R}$ is a reward function, and $\gamma \in [0,1]$ is a discount factor for weighting long-term rewards. Within this stochastic environment, the action selection strategy is determined by a deterministic policy, $\pi : \mathcal{S} \rightarrow \mathcal{A}$.

To assess the quality of a policy, the expected discounted sum of rewards when starting at state $\mathbf{s}$ and following policy $\pi$ can be computed using the value function $V^\pi : \mathcal{S} \rightarrow \mathbb{R}$. The value function starting from state $\mathbf{s}$ is defined as
\begin{align}
    V^\pi(\mathbf{s}) = \mathbb{E}_\pi\left[ \sum_{i=0}^{\infty} \gamma^{i} r_{i} \mid \mathbf{s}_{\text{init}}=\mathbf{s} \right] ,
\end{align}
where $r_i$ represents the reward at time $i$ and $\mathbf{s}_{\text{init}}$ is the initial state.

In control problems where we are interested in improving our action selection strategy, it is useful to consider the action-value function $Q^\pi : \mathcal{S} \times \mathcal{A} \rightarrow \mathbb{R}$. Given a policy $\pi$, $Q^\pi$ represents the expected discounted sum of rewards after taking action $a$ and state $\mathbf{s}$, and following policy $\pi$ thereafter, i.e.,
\begin{align}
   Q^\pi(\mathbf{s},a) = \mathbb{E}_\pi\left[ \sum_{i=0}^{\infty} \gamma^{i} r_{i} \mid \mathbf{s}_{\text{init}}=\mathbf{s}, a_{\text{init}} = a \right] 
\end{align}
Due to the Markovian property of our MDP, the action-value function (as well as the value function) is a fixed point of the Bellman operator $Q^\pi = \mathcal{T}_\pi Q^\pi$, where the operator $\mathcal{T}_\pi$ is defined as 
\begin{align}
    (\mathcal{T}_\pi Q)(\mathbf{s},a) = R(\mathbf{s},a) + \gamma \int_\mathcal{S} Q(\mathbf{s}', \pi(\mathbf{s}') )d\mathcal{P}(\mathbf{s}'|\mathbf{s},a)
\end{align}
or, equivalently in vector form, as $\mathcal{T}_\pi Q = \mathcal{R} + \gamma \mathcal{P}^\pi Q$, where $\mathcal{R} \in \mathbb{R}^{|\mathcal{S}||\mathcal{A}|}$ is a reward vector and $\mathcal{P}^\pi \in  \mathbb{R}^{|\mathcal{S}||\mathcal{A}| \times|\mathcal{S}||\mathcal{A}|}$ is the induced transition matrix when following policy $\pi$ after a next state transition according to $\mathcal{P}(\mathbf{s}'|\mathbf{s},a)$. 

For a given MDP, the optimal action-value function is defined as $Q^*(\mathbf{s},a) = \sup_\pi Q^\pi(\mathbf{s},a)$ for all states and actions $(\mathbf{s},a) \in \mathcal{S} \times \mathcal{A}$. For a given action-value function $Q$, we define a greedy policy $\pi$ as $\pi(\mathbf{s}) = \arg\max_{a\in\mathcal{A}} Q(\mathbf{s},a)$ for all $\mathbf{s} \in \mathcal{S}$. The greedy policy with respect to the optimal action-value function $Q^*$ is then an optimal policy, denoted as $\pi^*$. Hence, obtaining $Q^*$ allows us to arrive at an optimal action selection strategy. 

\subsection{Generalized Framework}
\label{Kernel-Sieve Hybrid}

For an arbitrary policy $\pi$, we introduce a generalized framework for modeling the action-value function $Q^\pi$ as a sum of nonparametric additive component functions. Our approach handles both discrete (i.e., $\mathcal{A} = \{1,\dots,k\}$) and continuous (i.e., $\mathcal{A} = \mathbb{R}$) action spaces, while allowing for the incorporation of potentially time-varying or time-invariant variables.

First, we present our generalized nonparametric framework for modeling $Q^\pi$ as
\begin{align}
    Q^\pi(\mathbf{s},a,x) &=  g_a(x) + \sum_{j=1}^df_{j,a}(\mathbf{s}_j,x) + \epsilon.
    \label{general-model}
\end{align}
Under this model, we expand the input space of $Q^\pi$ to include the candidate variable $x \in \mathbb{R}$, and discretize the action space such that $a \in \{1,\dots,k\}$ if $\mathcal{A}$ is not already discrete. Accordingly, $g_a(\cdot)$ represents the additive marginal effect of $x$ under action $a$, and $f_j(\cdot, \cdot)$ represents the additive joint effect of interactions between $x$ and state features $\mathbf{s}_j$ under action $a$. Without making specific assumptions on the functional form of $g_a(\cdot)$ and $f_{j,a}(\cdot, \cdot)$, our model allows us to carefully examine additive nonlinear relationships that exist among relevant state features, actions, and the variable $x$. 

Second, our choice in $x$ allows us to explore several unique representations of the additive components in \eqref{general-model}. For example, $x$ can represent time-varying or time-invariant confounders (e.g., age, gender, or the number of days since a surgical event) as well as continuous-valued actions $a \in \mathbb{R}$: 
\begin{align}
    x = \begin{cases} 
    \hspace{2pt}\mathbf{s}_0, \hspace{4pt} \text{i.e., a candidate state feature or confounder,}  \\
    \hspace{2pt}a, \hspace{8pt}\text{i.e., a continuous action}
    \end{cases}.
\end{align}

 \begin{example}
 \label{example-2.1}
    When $x = \mathbf{s}_0$, the additive functions in (\ref{general-model}) respectively equate to $g_a(x) = g_a(\mathbf{s}_0)$ and $f_{j,a}(\mathbf{s}_j,x) = f_{j,a}(\mathbf{s}_j,\mathbf{s}_0)$. Furthermore, we can augment the state space $S$ using $x$ to form $S_+ = \{x, s_1, \dots, s_d \}$, and succinctly represent $Q^\pi(\mathbf{s},a,\mathbf{s}_0)$ as $Q^\pi(\mathbf{s}_+,a)$, where $\mathbf{s}_+ \in S_+$ and 
    \begin{align}
        Q^\pi(\mathbf{s}_+,a) &= g_a(\mathbf{s}_0) + \sum_{j=1}^df_{j,a}(\mathbf{s}_j,\mathbf{s}_0) + \epsilon.
        \label{eq-2.1}
    \end{align}
     Thus, under the discrete action $a$, $g_a(\mathbf{s}_0)$ models the nonlinear marginal effect of the confounder or state feature $\mathbf{s}_0$, whereas $f_{j,a}(\mathbf{s}_j,\mathbf{s}_0)$ models the nonlinear interaction between $\mathbf{s}_0$ and state features $\mathbf{s}_j$. 
 \end{example}

\begin{example}
\label{example-2.2}
    Similarly, when $x = a$, the additive functions in (\ref{general-model}) respectively equate to $g_a(a) = g(a)$ and $f_{j,a}(\mathbf{s}_j,a) = f_j(\mathbf{s}_j,a)$. Under this choice of $x$, we avoid explicit discretization of the action space $\mathcal{A}$ and directly treat $a$ as a continuous action. Thus, for a given state action pair, $Q^\pi(\mathbf{s},a,a)$ reduces to $Q^\pi(\mathbf{s},a)$, where
     \begin{align}
        Q^\pi(\mathbf{s},a) &= g(a) + \sum_{j=1}^df_{j}(\mathbf{s}_j,a) + \epsilon,
        \label{eq-2.2}
    \end{align}
     and the additive marginal effect of selecting a continuous action $a$ is modeled as $g(a)$, and $f_j(\mathbf{s}_j,a)$ represents the additive effect of selecting action $a$ under state feature value $\mathbf{s}_j$.
\end{example} 
 
\section{Kernel Sieve Hybrid - Least Squares Policy Iteration}
\label{KSH-LSPI Section}
We introduce a general approach for estimating $Q^\pi(\mathbf{s},a)$ for both discrete and continuous action spaces. This estimation strategy provides an intuitive way to (1) locally approximate the action-value function as an additive model with independent state features spanned by a B-spline basis expansion, (2) retrieve an estimate of the nonlinear additive components, and (3) obtain a sparse representation of the action-value function by selecting relevant regions of the domain of the component functions.

\subsection{Basis Expansion}
First, we model the action-value function using a centered B-spline basis expansion of the additive component functions. Let $\{\psi_1,\dots, \psi_m \}$ be a set of normalized B-spline basis functions. For each component function, we project $f_{j,a}$ onto the space spanned by the basis, $\mathcal{B}_m = \text{Span}(\psi_1, \dots, \psi_m)$. Thus, $f_{j, a}(\mathbf{s}_j, x)$ can be expressed as $ \sum_{\ell=1}^m\varphi_{j\ell}(\mathbf{s}_j)\boldsymbol{\beta}_{j\ell;a}(x)$, 
where $\varphi_{j\ell}$ are locally centered B-spline basis functions defined as $\varphi_{j\ell}(\mathbf{s}) = \psi_\ell(\mathbf{s}) - E[\psi_\ell(\mathbf{s}_j)]$ for the $j$-th component function and the $\ell$-th basis component. 

As we will discuss in Section \ref{KSH-Weight}, our estimation strategy relies on performing a locally-weighted least-squares minimization of an objective criterion with respect to a fixed value of the variable $x$. As such, we locally express our model in $(\ref{general-model})$ by (i) setting $x=z$, where $z \in \mathcal{X}$ is some arbitrary fixed value, and by (ii) using the aforementioned centered B-spline basis expansion:
\begin{align}
    Q^\pi(\mathbf{s},a,x=z) \approx \alpha_{a,z} + \sum_{j=1}^d \sum_{\ell=1}^m \varphi_{j\ell}(\mathbf{s}_j)\beta_{j\ell;a,z} \quad .  \label{local-ksh-model}
\end{align}
Under this local model, $\alpha_{a,z} \in \mathbb{R}$ represents that marginal effect $g_a(x)$ when $z$ is a fixed value of $x$. Accordingly,  $\beta_{j\ell;a,z} \in \mathbb{R}$ is the coordinate corresponding to the $\ell$-th B-spline basis of the $j$-th state-feature under $z$. 

In Examples \ref{example-2.1} and \ref{example-2.2}, we observe two choices for representing $x$ that highlight the generalizability of our model structure. Under theses examples, the local additive components in (\ref{local-ksh-model}) can also be re-expressed as follows: 
\begin{align}
    \alpha_{a,z} = \begin{cases} 
    \hspace{2pt}\alpha_{a,z} \hspace{8pt} \text{when} \hspace{4pt} x = \mathbf{s}_0, \hspace{4pt}  \\
    \hspace{2pt}\alpha_{z} \hspace{15pt} \text{when} \hspace{4pt} x = a
    \end{cases} \quad \text{and} \quad 
    \beta_{j\ell;a,z} = \begin{cases} 
    \hspace{2pt}\beta_{j\ell;a,z} \hspace{8pt} \text{when} \hspace{4pt} x = \mathbf{s}_0, \hspace{4pt}  \\\hspace{2pt}\beta_{j\ell;z} \hspace{15pt} \text{when} \hspace{4pt} x = a
    \end{cases}.
\end{align}

Since the dynamics of the MDP are unknown, our estimation strategy relies on a batch dataset $\mathcal{D} = \{(\mathbf{s}^{[i]}, a^{[i]}, r^{[i]}, \mathbf{s}^{[i]}{'}, x^{[i]})\}_{i=1}^N$ of sampled transitions from the MDP of interest, where $\mathbf{s}^{[i]}{'} \sim P(\cdot|\mathbf{s}^{[i]},a^{[i]})$, and $x^{[i]}$ is the associated value of the candidate variable $x$. When we consider all observations in the dataset $\mathcal{D}$, we can equivalently re-express (\ref{local-ksh-model}) in vector form as
\begin{align}
    Q_{\boldsymbol{\beta}_+}^\pi = \tilde{\boldsymbol{\Phi}}\boldsymbol{\beta}_+ 
\end{align}
where $\boldsymbol{\beta_+} = (\boldsymbol{\beta}_{1+}^T,\dots, \boldsymbol{\beta}_{|\mathcal{A}|+}^T)^T$ $\in \mathbb{R}^{(1+dm)|\mathcal{A}|}$, $\boldsymbol{\beta}_{a+} = (\alpha_{a,z}, \boldsymbol{\beta}_{1;a,z}^T,\dots,\boldsymbol{\beta}_{d;a,z}^T)^T \in \mathbb{R}^{1+dm}$, and 
\begin{align}
\tilde{\boldsymbol{\Phi}}=\left(\hspace{0pt}\begin{array}{c}
\phi\left(\mathbf{s}^{[1]}, a^{[1]},\right)^T \\
\vdots \\
\phi\left(\mathbf{s}^{[N]}, a^{[N]},\right)^T
\end{array}\hspace{0pt}\right) = \left(\hspace{0pt}\begin{array}{c}
\varphi_+\left(\mathbf{s}^{[1]}\right)^T \mathbb{1}(a^{[1]}=1) \cdots \varphi_+\left(\mathbf{s}^{[1]}\right)^T \mathbb{1}(a^{[1]}=k)   \\
\vdots \\
\varphi_+\left(\mathbf{s}^{[N]}\right)^T \mathbb{1}(a^{[N]}=1) \cdots \varphi_+\left(\mathbf{s}^{[N]}\right)^T \mathbb{1}(a^{[N]}=k) 
\end{array}\hspace{0pt}\right)\label{vector-form}
\end{align} 
such that $\tilde{\boldsymbol{\Phi}} \in \mathbb{R}^{N \times (1+dm)|\mathcal{A}|}$, $\varphi_+(\mathbf{s}) = \big(1 \hspace{5pt} \varphi_2(\mathbf{s}_{1})\cdots\varphi_d(\mathbf{s}_{d})\big)^T \in \mathbb{R}^{1+dm}$ and $\varphi_j(\mathbf{s}_{j}) \in \mathbb{R}^m$ is a B-spline basis component vector. Note that $E[\psi_\ell(\mathbf{s}_j)]$ is estimated as $\Bar{\psi}_{j\ell} = N^{-1}\sum_{i=1}^N\psi_\ell(\mathbf{s}_{j}^{[i]})$ by using a sample of $N$ data points from the dataset $\mathcal{D}$.

\subsection{Kernel-Weighted Least Squares Fixed Point Approximation}
\label{KSH-Weight}

\begin{figure}[t]
    \centering
    \includegraphics[width=0.95\linewidth]{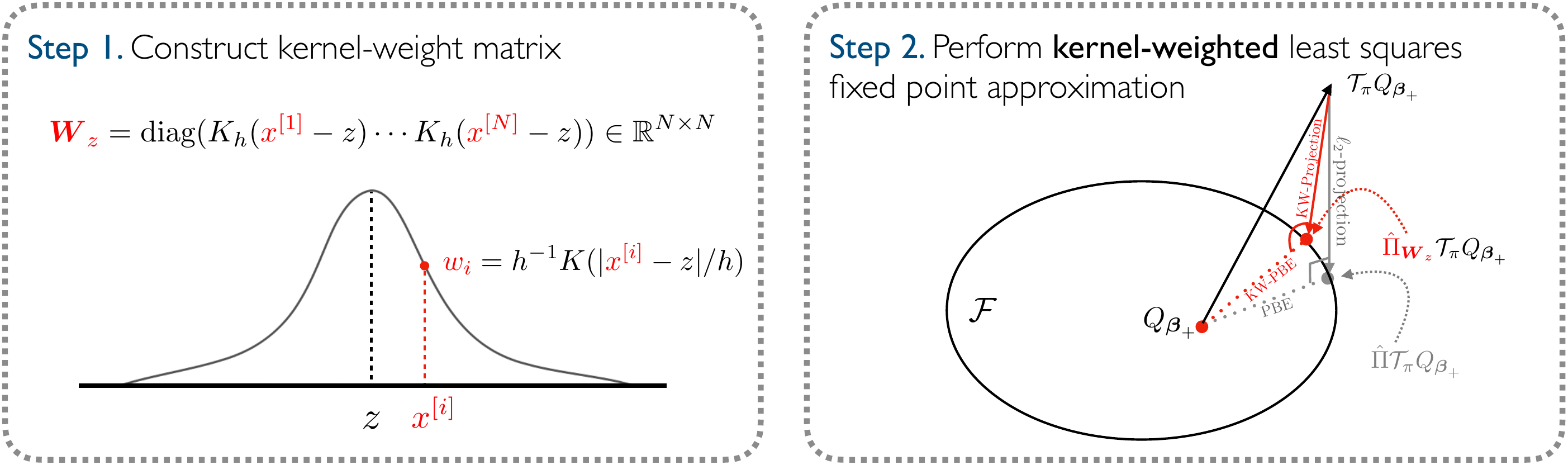}
    \caption[Long caption.]{\footnotesize\textit{A step-by-step illustration of kernel-weighted least squares fixed point approximation. First, using observations gathered in the batch dataset $\mathcal{D}$, we construct a diagonal kernel-weight matrix $\boldsymbol{W}_z$, where each diagonal weight is a function of the distance between the observed candidate variable $x^{[i]}$ and the fixed value $z$. Second, let $\mathcal{F}$ be $\mathcal{C}(\boldsymbol{\Phi})$, i.e., the space of approximate value functions. Since applying the Bellman operator $\mathcal{T}_\pi$ to an arbitrary value function $Q_{\boldsymbol{\beta}_+ }$ can push the resulting quantity $\mathcal{T}_\pi Q^\pi_{\boldsymbol{\beta}_+ }$ out of the space $\mathcal{F}$, we perform a projection step using the constructed kernel-weighted matrix $\boldsymbol{W}_z$ (detailed in \ref{KSH-Weight}). This approach differs from classical least squares fixed point approximation (shown in grey), where an $\ell_2$-projection operator $\hat{\Pi}$ is used. Lastly, we find $\hat{\boldsymbol{\beta}}_+$ that minimizes the $\ell_2$-norm between $Q_{\boldsymbol{\beta}_+ }$ and $\hat{\Pi}_{\boldsymbol{W}_z}\mathcal{T}_\pi Q^\pi_{\boldsymbol{\beta}_+ }$, i.e., the kernel-weighted projected bellman error.}}
    \label{fig:kernel_weight_approx}
\end{figure}

We estimate our model parameters $\boldsymbol{\beta}_+$ by minimizing a kernel-weighted version of the classical projected Bellman error ($\text{PBE}$). 

In LPSI, we observe that a simple procedure for estimating a linear action-value function is to force the approximate function to be a fixed point under the projected Bellman operator (i.e., $\Pi\mathcal{T}_\pi Q_{\boldsymbol{\beta}_+ } \approx Q_{\boldsymbol{\beta}_+ }$). 
For this condition to hold, the fixed point of the Bellman operator $\mathcal{T}_\pi$ must lie in the space of approximate value functions spanned by the basis functions over all possible state action pairs, $\mathcal{C}(\boldsymbol{\Phi})$. By construction, it is known that $Q_{\boldsymbol{\beta}_+ } = \boldsymbol{\Phi} \boldsymbol{\beta}_+  \in \mathcal{C}(\boldsymbol{\Phi})$. However, since there is no guarantee that $\mathcal{T}_\pi Q_{\boldsymbol{\beta}_+ }$ (i.e., the result of the Bellman operator) is in  $\mathcal{C}(\boldsymbol{\Phi})$, it first must be projected onto $\mathcal{C}(\boldsymbol{\Phi})$ using the projection operator $\Pi$, such that $\Pi \mathcal{T}_\pi Q ^\pi_{\boldsymbol{\beta}_+ } = \boldsymbol{\Phi} \boldsymbol{u}^*$ where $\boldsymbol{u}^*$ is the solution to the following least-squares problem
\begin{align}
    \boldsymbol{u}^* &= \argmin_{Q \in \mathcal{C}(\boldsymbol{\Phi})}\|Q_{\boldsymbol{\beta}_+ } - \mathcal{T}_\pi Q_{\boldsymbol{\beta}_+ } \|_2^2 = \argmin_{\boldsymbol{u} \in \mathbb{R}^k}\|\boldsymbol{\Phi} u - \mathcal{T}_\pi\boldsymbol{\Phi} \boldsymbol{\beta}_+ \|_2^2 . \label{l2_projection}
\end{align}
Empirically, $\boldsymbol{u}^*$ can be estimated using a sample-based feature design matrix $\tilde{\boldsymbol{\Phi}}$ constructed from a dataset of $N$ transitions $\mathcal{D}$, 
\begin{align}
        \boldsymbol{u}^* &= \argmin_{\boldsymbol{u} \in \mathbb{R}^k}\| \tilde{\boldsymbol{\Phi}} \boldsymbol{u} - \hat{\mathcal{T}}_\pi \tilde{\boldsymbol{\Phi}} \boldsymbol{\beta}_+ \|_2^2  \\
        &= \argmin_{\boldsymbol{u} \in \mathbb{R}^k}\sum_{i=1}^N\left(\phi(\mathbf{s}^{[i]},a^{[i]})^T \boldsymbol{u} - \Big[ r^{[i]} + \gamma  \phi(\mathbf{s}^{[i]}{'},\pi(\mathbf{s}^{[i]}{'}))^T\boldsymbol{\beta}_+ \Big] \right)^2
\end{align}
where $\hat{\mathcal{T}}_\pi$ is the empirical Bellman operator $(\hat{\mathcal{T}}_\pi Q_\beta)(\mathbf{s},a) = r(\mathbf{s},a) + \gamma \hspace{2pt} Q_\beta(\mathbf{s}',\pi(\mathbf{s}))$ defined using a single transition $\{\mathbf{s},a,r,\mathbf{s}'\}$ from $\mathcal{D}$. 

Rather than performing the projection step according to an $\ell_2$-norm, we propose using a \textit{kernel-weighted} norm with weights that are centered at a fixed value $z$ that lies within the domain of the candidate variable $x$. Let $K : \mathcal{X} \rightarrow \mathbb{R}$ be a symmetric kernel function with bounded support. We denote $K_h(\cdot) = h^{-1}K(\cdot/h)$, where $h > 0$ is the bandwidth. The solution $\boldsymbol{u}_z^*$ to the kernel-weighted projection step is estimated as follows
\begin{align}
        \boldsymbol{u}_z^* 
        &= \argmin_{\boldsymbol{u} \in \mathbb{R}^k}\sum_{i=1}^N K_h(x^{[i]} - z )\left(\phi(\mathbf{s}^{[i]},a^{[i]})^T \boldsymbol{u} - \Big[ r^{[i]} + \gamma  \phi(\mathbf{s}^{[i]}{'},\pi(\mathbf{s}^{[i]}{'}))^T\boldsymbol{\beta}_+ \Big]\right)^2 \\
        &= \argmin_{\boldsymbol{u} \in \mathbb{R}^k} \left( \tilde{\boldsymbol{\Phi}} \boldsymbol{u} - \hat{\mathcal{T}}_\pi \tilde{\boldsymbol{\Phi}} \boldsymbol{\beta}_+  \right)^T  \boldsymbol{W}_z \left( \tilde{\boldsymbol{\Phi}} \boldsymbol{u} - \hat{\mathcal{T}}_\pi \tilde{\boldsymbol{\Phi}} \boldsymbol{\beta}_+  \right) = (\tilde{\boldsymbol{\Phi}}^T \boldsymbol{W}_z \tilde{\boldsymbol{\Phi}})^{-1}\tilde{\boldsymbol{\Phi}}^T \boldsymbol{W}_z \hat{\mathcal{T}}_\pi\tilde{\boldsymbol{\Phi}} \boldsymbol{\beta}_+ ,
\end{align}
where $\boldsymbol{W}_z = \operatorname{diag}(K_h(x^{[1]}-z) \cdots K_h(x^{[N]}-z)) \in \mathbb{R}^{N \times N}$ is a diagonal kernel-weight matrix.  Under this weighted norm, transitions with a candidate variable $x^{[i]}$ that are local to $z$ contribute more to the overall fit of the least squares minimization. Accordingly, the empirical kernel-weighted projection operator is $\hat{\Pi}_{\boldsymbol{W}_z} = \tilde{\boldsymbol{\Phi}}(\tilde{\boldsymbol{\Phi}}^T \boldsymbol{W}_z \tilde{\boldsymbol{\Phi}})^{-1}\tilde{\boldsymbol{\Phi}}^T \boldsymbol{W}_z$.

Using the projection operator $\hat{\Pi}_{\boldsymbol{W}_z}$, we can now directly find $\boldsymbol{\beta_+}$ that minimizes the  kernel-weighted empirical PBE represented as
\begin{align}
    \mathcal{E}_\mathcal{D} = \|Q_{\boldsymbol{\beta}_+}^\pi - \hat{\Pi}_{\boldsymbol{W}_z} \mathcal{T}_\pi Q^\pi_ {\boldsymbol{\beta}_+}\|_2^2 = \|\tilde{\boldsymbol{\Phi}} \boldsymbol{\beta}_+ - \tilde{\boldsymbol{\Phi}}\underbrace{(\tilde{\boldsymbol{\Phi}}^T \boldsymbol{W}_z \tilde{\boldsymbol{\Phi}})^{-1}\tilde{\boldsymbol{\Phi}}^T \boldsymbol{W}_z \mathcal{T}_\pi\tilde{\boldsymbol{\Phi}} \boldsymbol{\beta}_+}_{\tilde{g}(\boldsymbol{\beta}_+)}\|_2^2  .
 \label{WEMSPBE}
\end{align}
Since $\tilde{\boldsymbol{\Phi}} \tilde{g}(\beta) \in \mathcal{C}(\tilde{\boldsymbol{\Phi}})$, minimizing this objective function is equivalent to solving for $\boldsymbol{\beta}_+$ in $\boldsymbol{\Phi} \boldsymbol{\beta}_+ = \boldsymbol{\Phi} \tilde{g}(\boldsymbol{\beta}_+)$, which can be simplified as 
\begin{align}
    \underbrace{\tilde{\boldsymbol{\Phi}}^T \boldsymbol{W}_z \big(\tilde{\boldsymbol{\Phi}} - \gamma\tilde{\boldsymbol{\Phi}}'\big)}_{\boldsymbol{A}_{z}}\boldsymbol{\beta}_+ = \underbrace{\tilde{\boldsymbol{\Phi}}^T \boldsymbol{W}_z\tilde{R}}_{\boldsymbol{b}_{z}}, 
    \label{result}
\end{align} 
where $\tilde{\boldsymbol{\Phi}}' = \left(\phi(\mathbf{s}^{[1]}{'},\pi(\mathbf{s}^{[1]}{'}))^T \hspace{2pt} \cdots \hspace{4pt} \phi(\mathbf{s}^{[N]}{'},\pi(\mathbf{s}^{[N]}{'}))^T \right)^T$. Thus, the solution to minimizing the kernel-weighted empirical PBE can be obtained analytically as $\hat{\boldsymbol{\beta}}_+ = \boldsymbol{A}_z^{-1}\boldsymbol{b}_z$. This procedure is summarized in Figure \ref{fig:kernel_weight_approx}.
\subsection{Component-wise Regularization via Group Lasso}
\label{group-lasso-section}
Since we are interested in obtaining a sparse representation of the elements in  $\boldsymbol{\beta}_+$, we apply a penalty to an estimating equation $\mathcal{L}_z(\boldsymbol{\beta}_+)$ of (\ref{result}). Since the components of our basis functions are grouped by features, we incorporate a group Lasso penalty that performs group-level variable selection by jointly constraining all coefficients that belong to a given feature. Consequently, the primary objective function for our estimator is
\begin{align}
    \mathcal{L}_z(\boldsymbol{\beta}_+) + \lambda \mathcal{R}(\boldsymbol{\beta}_+) = -\frac{1}{2} \boldsymbol{\beta}_+^T \tilde{\boldsymbol{A}}_{z} \boldsymbol{\beta}_+ - \boldsymbol{\beta}_+^T \tilde{\boldsymbol{b}}_{z} 
    + \lambda \sum_a^{|\mathcal{A}|}\Big(\sqrt{m} \cdot|\alpha_a|+\sum_{j\geq2}\left\|\boldsymbol{\beta}_{j;a}\right\|_{2}\Big) ,
    \label{objective}
\end{align}
where $\lambda$ is a regularization parameter. Note that the group Lasso penalty $\mathcal{R}(\boldsymbol{\beta}_+)$ includes a $\sqrt{m}$ factor, which is used to appropriately scale the strength of the regularization term $\lambda$ applied to $|\alpha_a|$ and to that of the coefficients of the B-splines basis functions. This ensures that the grouped coefficients get evenly penalized. \\
\begin{algorithm}[t]
     \caption{{KSH-LSTDQ (via Randomized Coordinate Descent for Group Lasso)}} 
   \label{KSH-LSTD}
    \SetAlgoLined
    \textbf{Input:} $z$ (Fixed value), $\boldsymbol{\beta}^{(0)}_{+}$ (Initial weights), 
    $K(\cdot)$ (Kernel function), $0 \leq \gamma < 1$ (Discount factor), 
    $\mu$ (Step size), $\epsilon$ (Stopping Criteria), $\lambda$ (Regularization Parameter), $\pi$ (Current Policy) \\[0.45em]
    \textbf{Data:} Dataset of transitions $\mathcal{D} = \{(\mathbf{s}^{[i]}, a^{[i]}, r^{[i]}, \mathbf{s}^{[i]}{'}, x^{[i]})\}_{i=1}^N$ \\[0.45em]
    \textbf{Initialization:} 
    Construct $\boldsymbol{W}_z$, $\boldsymbol{\Phi}$ and $\boldsymbol{\Phi}'$\\
    \While{$||\boldsymbol{\beta}^{(t+1)}_+ - \boldsymbol{\beta}^{(t)}_+|| \geq \epsilon$}{
      Select $j \in [d]$ with probability $1/d$\\
      \For{$a \in \mathcal{A}$}{
       Update $\boldsymbol{\beta}^{(t+1)}_{j,a} \leftarrow \mathcal{U}_{\lambda_{j} / \mu}\left(\boldsymbol{\beta}_{j;a}^{(t)}-\mu \hspace{2pt} \boldsymbol{\Phi}^T_{\bigcdot j;a} \boldsymbol{W}_z \left(  \big(\boldsymbol{\Phi} - \gamma\boldsymbol{\Phi}'\big)\boldsymbol{\beta}_+^{(t)} - \tilde{R} \right) \right)$
     }
   }
\end{algorithm}
\indent To estimate $\boldsymbol{\beta}_+$ under the objective function \ref{objective}, we use the randomized coordinate descent method for composite functions proposed in \citet{Richtarik2014IterationFunction}. Under this procedure, we (1) randomly select a coordinate $j$ from $\{1,\dots,d\}$ under a fixed action $a$, and (2) update the current estimate of $\beta_{j;a}^{(t)}$. We then repeat steps (1) and (2)  until convergence to $\hat{\boldsymbol{\beta}}_+$. Each update in (2) can be written in closed form as
 \begin{align}
    U\left(\boldsymbol{\beta}_{j;a}^{(t)}\right)&=\mathcal{U}_{\lambda_{j} / \mu}\left(\boldsymbol{\beta}_{j;a}^{(t)}-\mu\nabla_{j;a} \mathcal{L}_z(\boldsymbol{\beta}_+^{(t)})\right)  \\[0.5em] 
    &=\mathcal{U}_{\lambda_{j} / \mu}\left(\boldsymbol{\beta}_{j;a}^{(t)}-\mu \hspace{2pt} \boldsymbol{\Phi}^T_{\bigcdot j;a} \boldsymbol{W}_z \left(  \big(\boldsymbol{\Phi} - \gamma\boldsymbol{\Phi}'\big)\boldsymbol{\beta}_+^{(t)} - \tilde{R} \right) \right)  \label{grad}
\end{align}
where $\mathcal{U}_\lambda$ is a soft-thresholding operator defined as $\mathcal{U}_\lambda(\mathbf{v})=\left(\mathbf{v} /\|\mathbf{v}\|_{2}\right) \cdot \max \left\{0,\|\mathbf{v}\|_{2}-\lambda\right\}$, $\mu$ is the step size, and $\lambda_j$ are regularization parameters. Note that $\lambda_1 = \lambda\sqrt{m}$ and $\lambda_j = \lambda$ $\forall j \in [d]$. Details of the estimation procedure of KSH-LSTDQ are provided in Algorithm \ref{KSH-LSTD}. \\
\indent This algorithm allows us to retrieve an estimate of $\boldsymbol{\beta}_+$ relative to the fixed value $z$ and, accordingly, approximate the additive functions in \ref{general-model}, as 
\begin{align}
    \hat{g}_a(z) = \hat{\alpha}_{a,z} \quad \text{and} \quad 
    \hat{f}_{j,a}(\mathbf{s}_j, z) = \sum_{k=1}^m \varphi_{jk}(\mathbf{s}_j)\hat{\beta}_{jk;a,z} \quad \forall j \geq 2.
\end{align}
To retrieve a nonlinear, smooth estimate of $g_a(\cdot)$ and $f_{j,a}(\mathbf{s}_j,\cdot)$ respectively, we compute the estimators $\hat{\alpha}_{a,z}$ and $\hat{f}_{j,a}(\mathbf{s}_j, z)$ for each value of $z$ contained within the set $\mathcal{Z} = \{z_1, \dots, z_{M}\}$ that densely covers the domain of $x$. This procedure amounts to running Algorithm $\ref{KSH-LSTD}$ $M$ times (i.e., once for each element in $\mathcal{Z}$).
\subsection{Approximate Policy Iteration}
\label{approximate_pi}
The aforementioned estimation strategy is a policy evaluation method for obtaining an approximate representation of the action-value function $Q^\pi$ under a fixed policy $\pi$. By using policy iteration, we can construct a procedure for estimating $Q^{ *}$ under an improved, or potentially optimal policy $\pi^*$ \citep{Howard1960DynamicProcesses, Bertsekas2011ApproximateMethods}.

\begin{algorithm}[t]
     \caption{{KSH-LSPI}} 
    \label{KSH-LSPI}
    \SetAlgoLined
    \textbf{Data:} 
    $\mathcal{D} = \{(\mathbf{s}^{[i]}, a^{[i]}, r^{[i]}, \mathbf{s}^{[i]}{'}, x^{[i]})\}_{i=1}^N$  \\[0.1em]
    \textbf{Input:} $\mathcal{Z}$ (Array of fixed values), $\{\boldsymbol{\beta}^{\text{init}}_{+(i)}\}_{i=1}^{|\mathcal{Z}|}$ (Initial weights), 
    $K(\cdot)$ (Kernel function), $0 \leq \gamma < 1$ (Discount factor), 
    $\mu$ (Step size), $\epsilon$ (Stopping Criteria), $\lambda$ (Regularization Parameter) \\[0.3em]
    \textbf{Initialization:} Construct $\boldsymbol{\Phi}$ \\[0.2em]
    $\mathcal{B}^{(t)} = [\boldsymbol{\beta}^{\text{init}}_{+(i)}]_{i=1}^{|\mathcal{Z}|}$ (Matrix of model weights) \\[0.3em]
    \While{$\|\mathcal{B}^{(t+1)}  - \mathcal{B}^{(t)}\|_F \leq \epsilon$}{
    Construct $\boldsymbol{\Phi}'$ w.r.t to a greedy policy $\pi(s) = \arg\max_a \hat{Q}^\pi(s,a)$ using (\ref{approximate_pi}). \\
        \For{$i \in \{1\dots, |\mathcal{Z}| \}$}{
           $\mathcal{B}_{i\hspace{1pt}\bigcdot}^{(t+1)} \leftarrow $ KSH-LSTDQ($\boldsymbol{\Phi}$, $\boldsymbol{\Phi}'$, $\mathcal{B}_{i\hspace{1pt}\bigcdot}^{(t)}$, $z_i$,\hspace{2pt}$\dots$)
        }
     }{}
\end{algorithm}

To perform policy iteration, we begin with an arbitrary policy $\pi_0$, or the behavioral policy $\pi_\text{b}$ used to generate $\mathcal{D}$. At each iteration $t$, we evaluate the current policy $\pi_t$ by estimating $Q^{\pi_t}$ according to \eqref{local-ksh-model} over a grid of local points $\mathcal{Z} = \{ z_1,\dots,z_M\}$. The policy improvement step follows by using the recently approximated action-value function $Q^{\pi_t}$ to generate the new greedy policy $\pi_{t+1}$. Since $Q^{\pi_t}$ is represented using a grid of $m$ local models (each computed with respect to each fixed value of $x$), the action selection strategy and representation of $\pi_{t+1}$ is closely determined by our choice of $x$. This process, as detailed in Algorithm \ref{KSH-LSPI}, repeats until convergence.  

\begin{example} 
\label{policy_iteration_discrete_action}
When $x=\mathbf{s}_0$, we represent the greedy policy $\pi(\mathbf{s})$ using the local model whose value of $z$ is closest to $\mathbf{s}_0$. In other words, let $\mathcal{Z} = \{z_1, \dots, z_{M}\}$ and $\mathcal{B} \in \mathbb{R}^{M \times (1+(d-1)m)|\mathcal{A}|}$ be a matrix of weights, where each row $i$ corresponds to a set of model weights estimated under the value $z_i$. The greedy policy $\pi(\mathbf{s})$ is defined as $\pi(\mathbf{s}) = \arg\max_a \phi\left(\mathbf{s},a\right)^T\mathcal{B}_{i^*\bigcdot}$, where $i^* = \arg\min_{i \in \{1\dots,M \}} |\mathbf{s}_0 - z_i|$.
\end{example} 

\begin{example}
When $x=a$, the greedy policy is represented as the fixed value $z$ of local model that maximizes its associated action-value function. Let $\mathcal{Z} = \{z_1, \dots, z_{M}\}$ and $\mathcal{B} \in \mathbb{R}^{M \times (1+(d-1)m)}$ be a matrix of weights, where each row $i$ corresponds to a set of model weights estimated under the value $z_i$. The greedy policy $\pi(\mathbf{s})$ is defined as $\pi(\mathbf{s}) = z_{i^*}$ where $i^* = \arg\max_i \phi\left(\mathbf{s},\cdot\right)^T\mathcal{B}_{i\bigcdot}$\hspace{2pt}.
\end{example}

\section{Simulation Study}
\label{simulation-study}

In this section, we perform a simulation study to examine the key properties of the KSH-LSTDQ and KSH-LSPI algorithms. We highlight the performance of the KSH-LSTDQ algorithm in estimating nonlinear marginal additive functions $g_a(x)$, examine its sensitivity to different environmental and model configurations, and compare the performance of the KSH-LSPI algorithm against several neural network-based approaches.  

\begin{figure}[!t]
    \centering
\begin{center}
    {\small $a=1$} \textcolor{Orange}{\threeemdash} \hspace{0.5em} {\small $a=0$} \textcolor{MidnightBlue}{\threeemdash}
\vspace{-1em}
\end{center}
\vspace{-0.5em}
    
    \subfloat[\centering Marginal effect $\hat{g}_a(\mathbf{s}_1)$ vs. Monte Carlo estimate of $U_{1}(\mathbf{s}_1,a)$.]{
\begin{tabular}{cccc}
  \scriptsize\rotatebox{90}{\hspace{35pt}$\text{Component Value}$}& \hspace{-15.5pt}
  \includegraphics[width=0.35\linewidth]
  {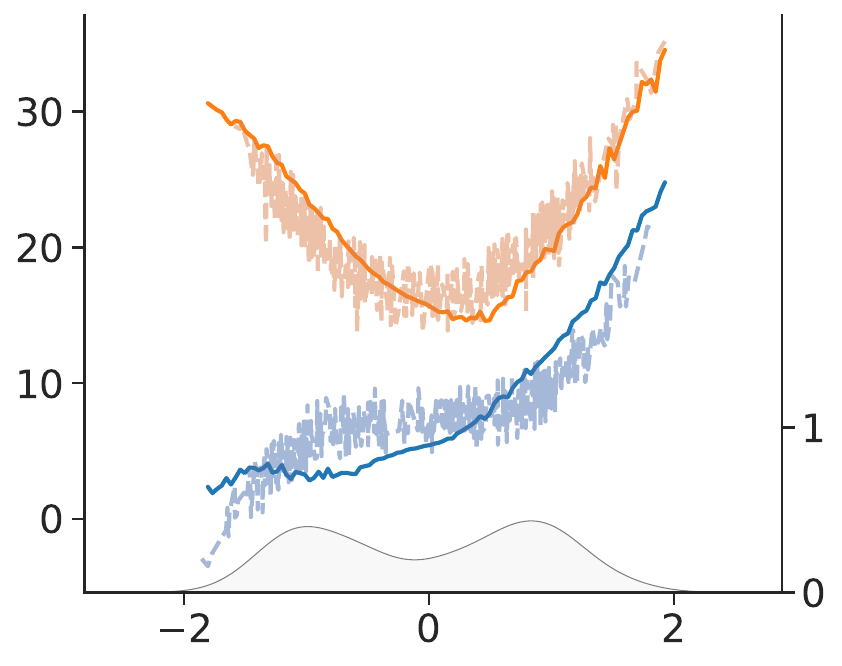} & \hspace{-17pt}
  \includegraphics[width=0.35\linewidth]{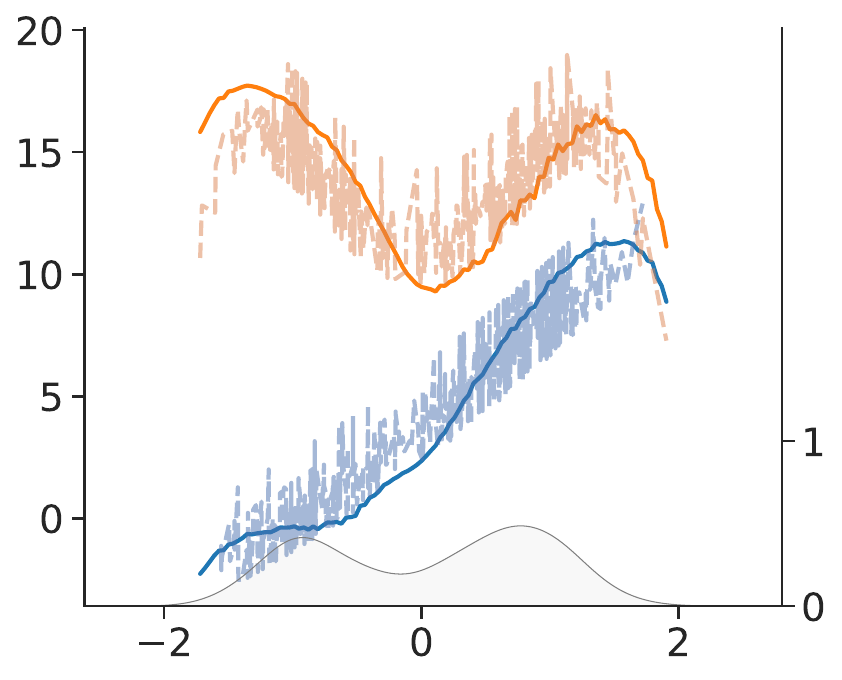}
  & {\hspace{-10pt}\scriptsize\rotatebox{90}{\hspace{53pt}$\text{Density}$}} \vspace{-0.75em} \\
  & \hspace{-7pt}\small$\mathbf{s}_1$ & 
  \hspace{-7pt}\small$\mathbf{s}_1$ 
\end{tabular}}
\hspace{-1.6em}
\subfloat[\centering Marginal effect $\hat{g}_a(\mathbf{s}_2)$ vs. Monte-Carlo estimate of $U_{2}(\mathbf{s}_2,a)$.]{
\begin{tabular}{cccc}
  \scriptsize\rotatebox{90}{\hspace{35pt}$\text{Component Value}$}& \hspace{-15.5pt} \includegraphics[width=0.35\linewidth]{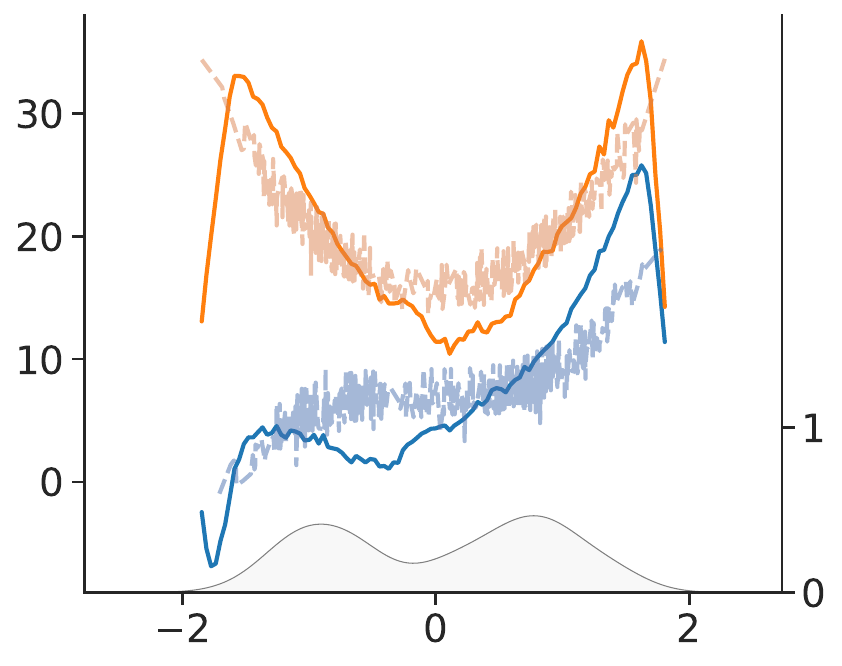} & \hspace{-17pt}
  \includegraphics[width=0.35\linewidth]{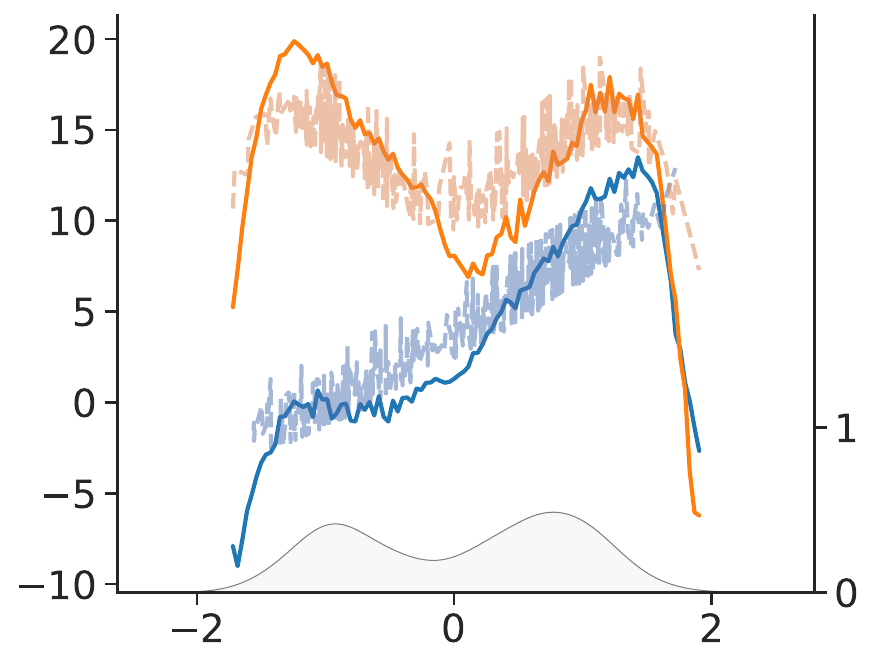} & {\hspace{-10pt}\scriptsize\rotatebox{90}{\hspace{53pt}$\text{Density}$}} \vspace{-0.75em}\\
  & \hspace{-9pt}\small$\mathbf{s}_2$ & 
  \hspace{-10pt}\small$\mathbf{s}_2$ 
\end{tabular}}
    \caption{\footnotesize\textit{A comparison of estimated marginal component functions $\hat{g}_a(\mathbf{s}_i)$ and MC-estimates of $u_i(\mathbf{s}_i,a)$ as described in Section \ref{simulation-study-marginal}. For each action, the solid lines represent the estimate of the marginal component function, $g_a(\mathbf{s}_i)$, of $Q^\pi(s,a)$ as modeled in Equation \ref{simulation-additive-model} under a bandwidth of $h=0.01$ (left) and $h=0.1$ (right) under $\gamma=0.5$, while the dashed line represents the Monte Carlo estimate of $U_i(\mathbf{s}_i,a)$. The observed distribution of the state feature $\mathbf{s}_i$ is displayed using the density in grey.  }}
    \label{fig:simulation_results}
  \end{figure}

\subsection{Estimating Marginal Components}
\label{simulation-study-marginal}

We consider a multidimensional, continuous state MDP with binary actions and an additive reward function. For each sampled trajectory, the initial state vector $\mathbf{s}^{(0)} \in \mathbb{R}^d$ is sampled uniformly from $[0, 1]^d$. At each time step $t$, we randomly sample an action $a^{(t)} \in \{0,1\}$ with probability $\frac{1}{2}$. Accordingly, the state transition function is defined as:
\begin{align}
    \mathbf{s}^{(t+1)} = (-1)^a (\sin(\mathbf{x} + a\mathbf{u}) + 0.1(\mathbf{s}^{(t)})^2) + \boldsymbol{\epsilon}
\end{align}
where $\mathbf{x} \sim \text{Unif}(0, 2)^d$, $\mathbf{u} \sim \text{Unif}(0, 1)^d$, and $\boldsymbol{\epsilon} \sim \mathcal{N}(0, \sigma^2\mathbf{I})$. This transition function allows us to explore complex, nonlinear dynamics into the MDP. We construct a reward function 
\begin{align}
    r(\mathbf{s},a) = u_1(\mathbf{s}_1,a) + u_2(\mathbf{s}_2,a)
    \label{additive-reward}
\end{align}
with reward components that are reliant only on the state features $\mathbf{s}_1$ and $\mathbf{s}_2$, where 
\begin{align}
    u_1(\mathbf{s}_1,a) &= (5\mathbf{s}_1^2 + 5)  \mathbb{1}(a=1) - (2\mathbf{s}_1^3 - 5)  \mathbb{1}(a=0), \\
    u_2(\mathbf{s}_2,a) &= (5\sin(\mathbf{s}_2^2) + 5) \mathbb{1}(a=1) + (4\mathbf{s}_2 - 5)  \mathbb{1}(a=0) 
\end{align}
and $g_j(\mathbf{s}_j,a) = 0 \hspace{3.5pt} \forall j \geq 3$.
For an arbitrary policy $\pi$, the construction of this reward function induces a corresponding action-value function that is additive with respect each non-zero reward component, specifically 
\begin{align*}
       Q^\pi(\mathbf{s},a) &= \mathbb{E}_\pi\left[ \sum_{i=0}^{\infty} \gamma^i r(\mathbf{s}^{(i)},a^{(i)}) \mid \mathbf{s}^{(0)}=\mathbf{s}, a^{(0)} = a \right] \\ &= \sum_{j=1}^2\mathbb{E}_\pi\left[ \sum_{i=0}^{\infty} \gamma^{i} u_j(\mathbf{s}_j^{(i)},a^{(i)}) \mid s^{(0)}=\mathbf{s}, a^{(0)} = a \right] \\
       &= U_1(\mathbf{s}_1,a) +  U_2(\mathbf{s}_2,a).
\end{align*}

\begin{figure}[!t]
    \centering    
    \subfloat[\centering Simulation results using a B-spline basis expansion.]{
\begin{tabular}{ccccc}
& \hspace{-7pt} $\dim{\mathbf{s}_i}$ & 
  \hspace{-12pt} $\lambda$ & \hspace{-11pt} $h$ & \hspace{-2pt} $\mu$ \\
  \scriptsize\rotatebox{90}{\hspace{9pt}$\text{Avg. Bellman Loss}$}& \hspace{-14.5pt}
  \includegraphics[width=0.22\linewidth]{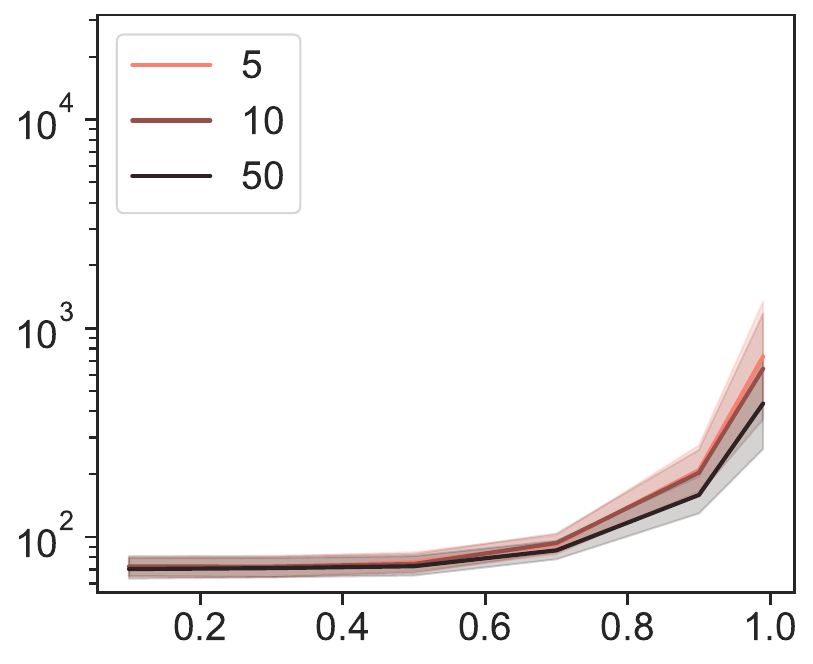} & \hspace{-16pt}
  \includegraphics[width=0.22\linewidth]{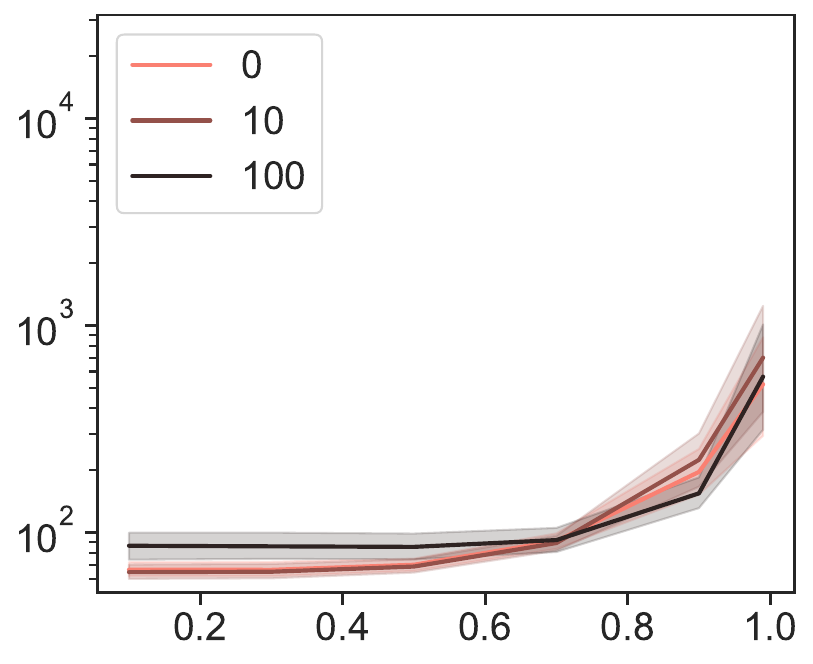}
  &  \hspace{-16pt}  \includegraphics[width=0.22\linewidth]{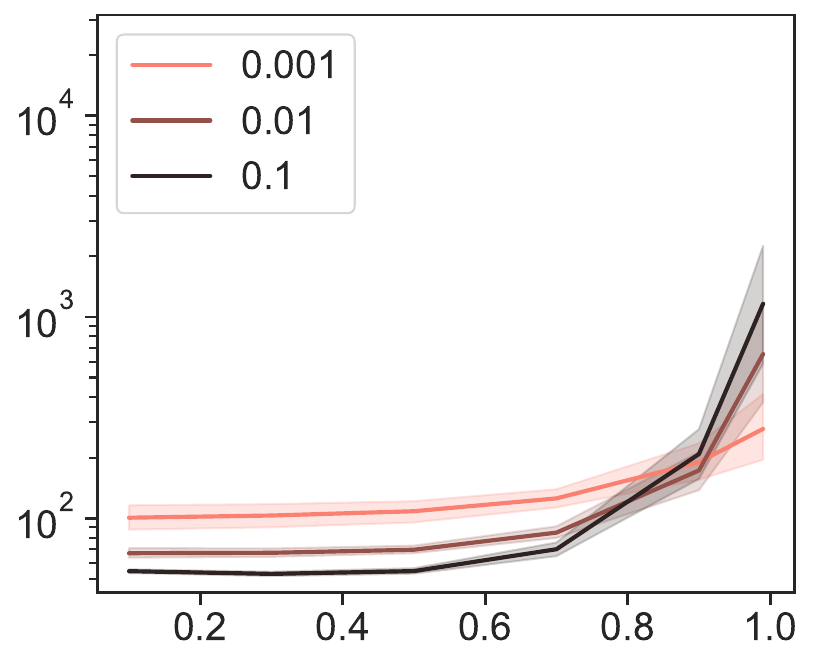} & \hspace{-8pt}
  \includegraphics[width=0.22\linewidth]{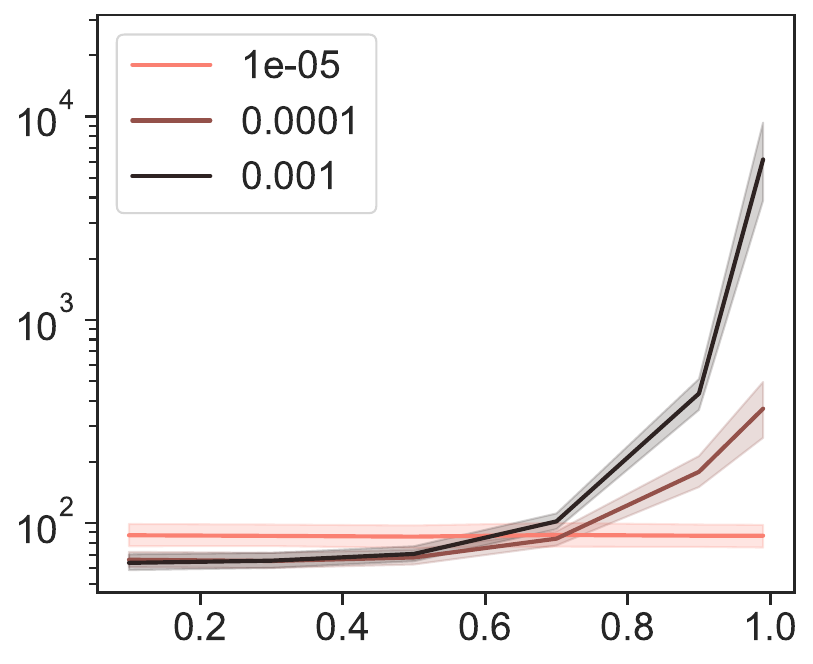}\\
   &  &   \hspace{93pt} $\gamma$
  & & 
\end{tabular}}
\hspace{-1.6em}
\subfloat[\centering Simulation results using a trigonometric polynomial basis expansion.]{
\begin{tabular}{ccccc}
& \hspace{-7pt} $\dim{\mathbf{s}_i}$ & 
  \hspace{-12pt} $\lambda$ & \hspace{-11pt} $h$ & \hspace{-2pt} $\mu$ \\
  \scriptsize\rotatebox{90}{\hspace{9pt}$\text{Avg. Bellman Loss}$}& \hspace{-14.5pt}
  \includegraphics[width=0.22\linewidth]{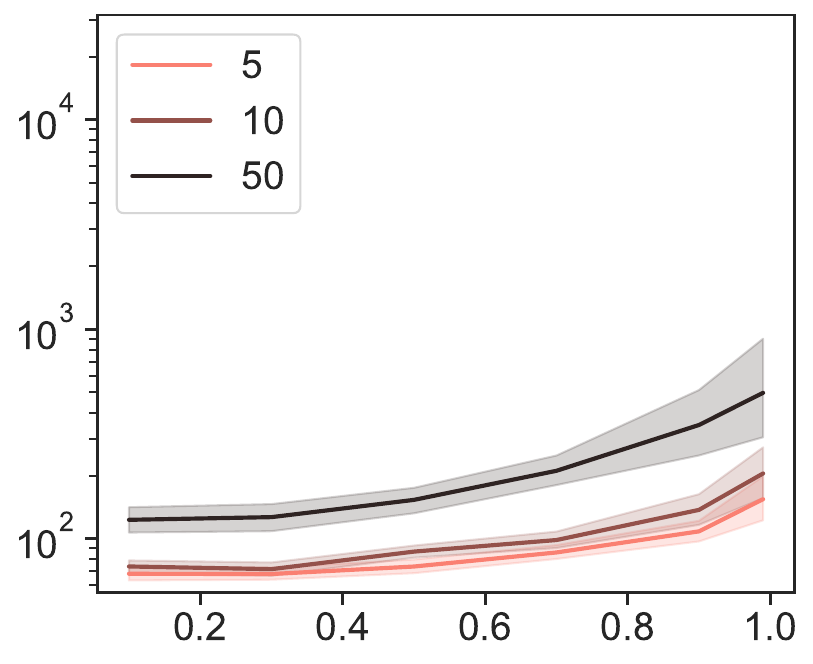} & \hspace{-16pt}
  \includegraphics[width=0.22\linewidth]{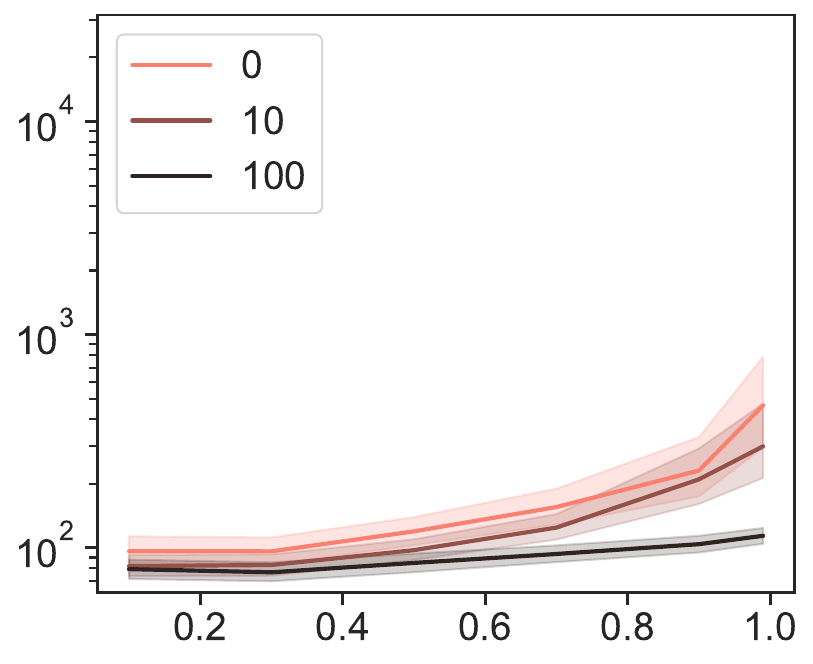}
  &  \hspace{-16pt}  \includegraphics[width=0.22\linewidth]{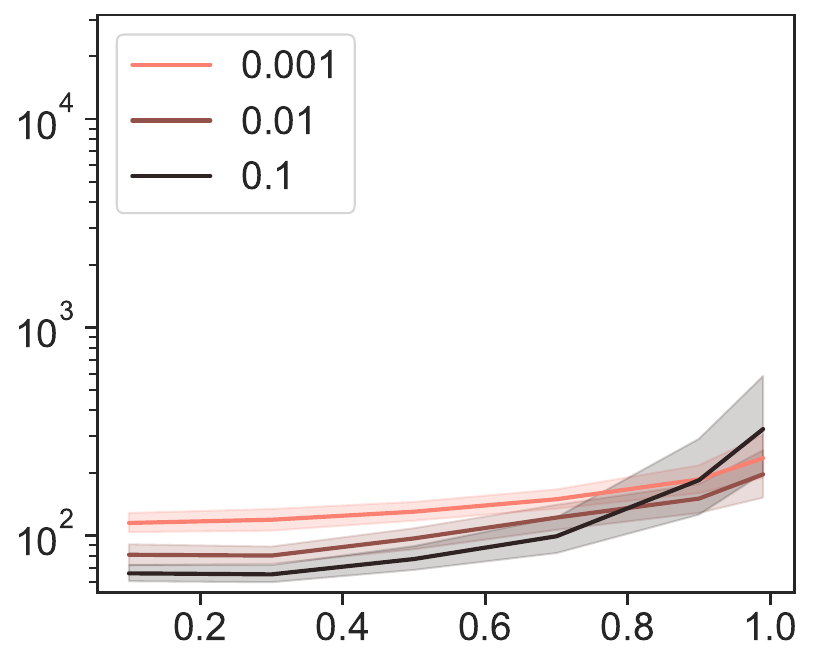} & \hspace{-8pt}
  \includegraphics[width=0.22\linewidth]{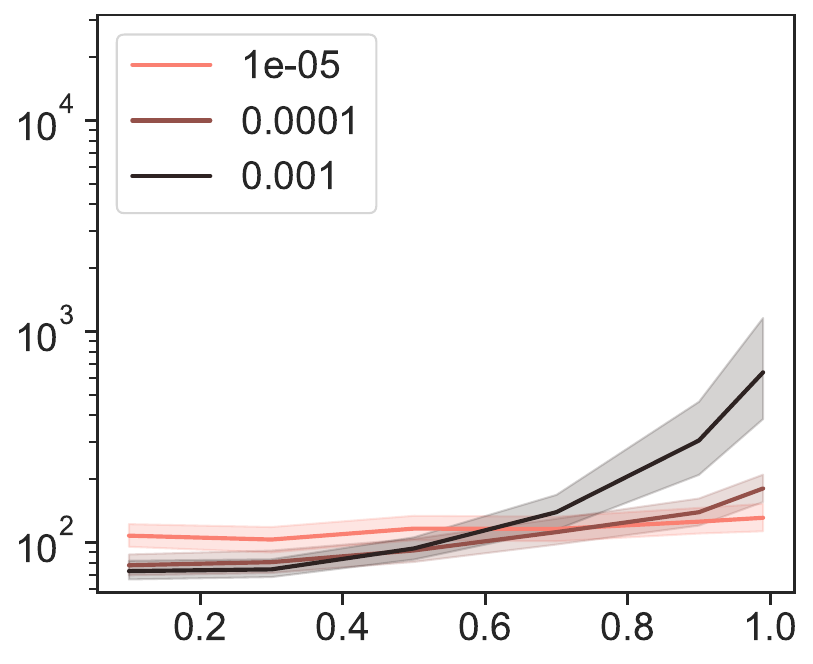}\\
   &  &   \hspace{93pt} $\gamma$
  & & 
\end{tabular}}
    \caption{\footnotesize\textit{Results of the simulation study described in Section \ref{simulation-study}. Average Bellman loss (\textit{y-axis}) vs discount factor $\gamma$ (\textit{y-axis}) stratified by the dimensionality of $\mathbf{s}_i$, regularization penalty $\lambda$, bandwidth $h$, and step size $\mu$. For each parameter setting, the solid lines represent the average Bellman loss as a function of $\gamma$, while the different line colors denote variations in $\dim \mathbf{s}_i$, $\lambda$, $h$, and $\mu$.}}
    \label{fig:simulation_results-ablation}
  \end{figure}

\indent Using $n$ trajectories sampled from this MDP (represented as a batch dataset $\mathcal{D}$), we evaluate the behavioral policy (i.e., $\pi(s^{[i]}) = a^{[i]}$) and retrieve the marginal component function $g_a(x)$ of the following nonparametric additive model
\begin{align}
    Q^\pi(\mathbf{s},a) &= g_a(\mathbf{s}_i) + \sum_{j\in [d]/i} f_{j,a}(\mathbf{s}_j,\mathbf{s}_i) + \epsilon,
    \label{simulation-additive-model}
\end{align}
where $x=\mathbf{s}_i$ and $i \in \{1,2\}$.  For each component function, we explored using a B-spline basis (with hyperparameters including bandwidth $h$, number of basis functions $m$, and the degree of each piece-wise polynomial) or a trigonometric basis function (hyperparameters include the number of basis functions $m$). We select the model's hyperparameters (including the regularization penalty $\lambda$ and step size $\mu$) using 5-fold cross-validation. For each combination of hyperparameters, we compute the average out-of-sample Bellman loss across the five folds and select hyperparameters $\theta^*$ that minimize the average loss. To measure the performance of our model against a target, we utilize Monte Carlo (MC) sampling on the MDP to retrieve a direct estimate of the $Q^{\pi}(s,a)$, evaluated as 
\begin{align}
\hat{Q}_\text{MC}^{\pi}(\mathbf{s},a) = \frac{1}{n}\sum_{i=1}^n\sum_{j=1}^\ell\gamma^{j}r_{ij},
\end{align}
where $\ell$ is the length of each trajectory and $n$ is the number of sampled trajectories. Since our action-value function is additive, we can similarly construct MC estimates for the component functions $U_1(\mathbf{s}_1,a)$ and $U_2(\mathbf{s}_2,a)$. Lastly, using a pre-specified grid of points $\mathcal{Z} =\{z_1, \dots, z_{M}\}$, we repeat Algorithm \ref{KSH-LSTD} $M$ times to obtain smooth estimates of $g_a(\mathbf{s}_i)$ as described in Section \ref{group-lasso-section}.

Figures \ref{fig:simulation_results} and \ref{fig:simulation_results-ablation} present the nonlinear marginal component functions of the estimated model, as described in Equation \eqref{simulation-additive-model}, and an ablation study over MDP and model hyperparameters, respectively. The dataset $\mathcal{D}$ comprised $n=100$ sampled trajectories, each with a length of $\ell=10$. MC estimates were obtained by sampling 100 trajectories of length $\ell$.

Figure \ref{fig:simulation_results} compares the MC estimate of $u_i(\mathbf{s}_i,a)$ with the estimated component function $\hat{g_a}(\mathbf{s}_i)$ derived from the KSH-LSTDQ estimator, using bandwidths of $h=0.01$ and $h=0.1$. In this scenario, we set the state space dimensionality to $d=10$ and the discount factor to $\gamma = 0.5$. As the kernel function's bandwidth increases, the model generates a smoother component function, attributable to larger weights being assigned to observations further from each local point $z$. Notably, in sparsely sampled regions of the domain, the component function values tend towards zero, particularly with smaller bandwidths. This effect is partially driven by the group Lasso penalty, which shrinks estimates within these sparse regions towards 0. While our model generally captures the underlying function's shape in non-sparse domain regions, our estimates may exhibit slight bias for complex functions.

Figure \ref{fig:simulation_results-ablation} illustrates results from an ablation study, depicting the average Bellman loss estimated across all folds on the y-axis against six different discount factors ($\gamma = 0.1,\dots,0.99$) on the x-axis. The results are further stratified by state space dimensionality $\dim \mathbf{s}_i$, regularization parameter $\lambda$, bandwidth $h$, step size $\mu$, and choice of basis expansion method.

Generally, we observe that as the discount factor $\gamma$ increases, the average Bellman loss decreases. This can be attributed to increased instability in model estimation, as future next-state transitions and actions (i.e., $\gamma\boldsymbol{\Phi}'$) are weighted more heavily, potentially leading to a singular matrix $\boldsymbol{A}_{z}$ in Equation \ref{result}. The results also indicate that the effect of increasing $\gamma$ can be mitigated by increasing the strength of the regularization parameter (especially in trigonometric polynomial basis functions) and by using smaller step sizes. Furthermore, we observe minimal sensitivity to changes in the state dimension for B-spline basis functions compared to trigonometric basis functions, where the average Bellman loss decreases as the state dimensionality grows. Lastly, we generally observe that as the bandwidth increases, the average Bellman loss across $\gamma$ decreases.

\begin{figure}[!t]
    \centering
    
    \subfloat[\centering Dimensionality of $\mathbf{s}_i$ is 5.]{
    \begin{tabular}{cc}
\scriptsize\rotatebox{90}{\hspace{50pt}$\text{Mean Regret}$} & \hspace{-10pt}\includegraphics[width=0.35\linewidth]
  {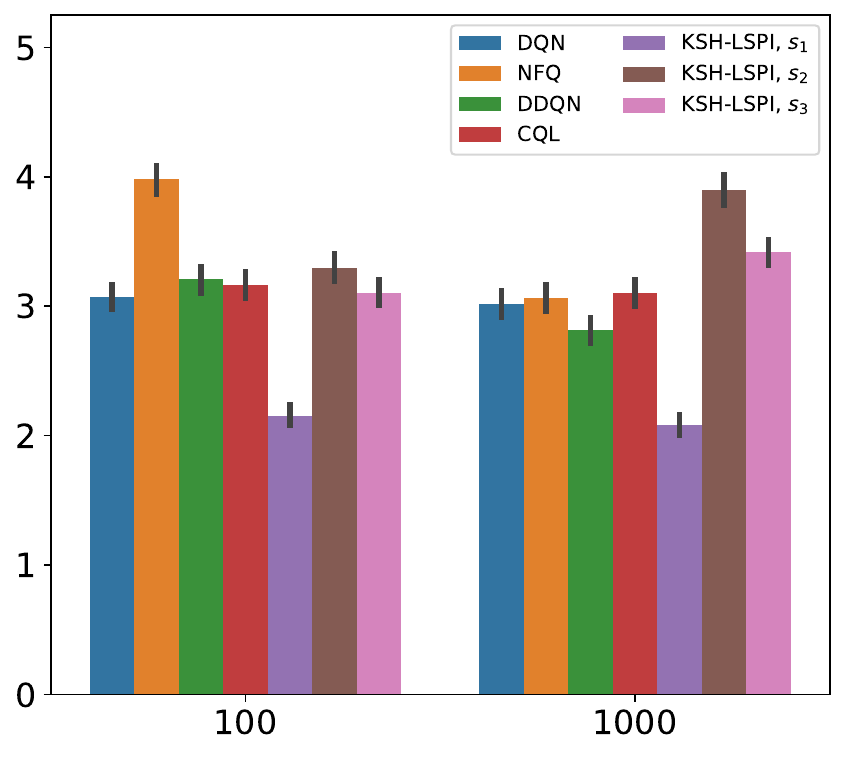} \vspace{-0.7em}\\
  & \hspace{-7pt} \scriptsize{Number of Episodes}
    \end{tabular}}
  \hspace{1em}
    \subfloat[\centering Dimensionality of $\mathbf{s}_i$ is 50.]{
    \begin{tabular}{cc}
\scriptsize\rotatebox{90}{\hspace{50pt}$\text{Mean Regret}$} & \hspace{-10pt}\includegraphics[width=0.35\linewidth]
  {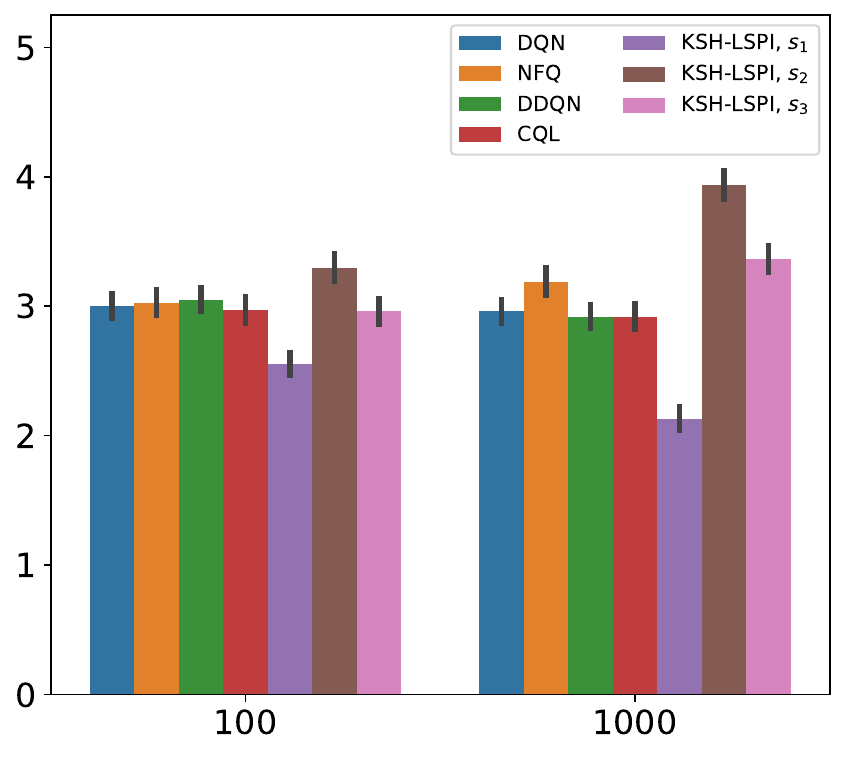} \vspace{-0.7em}\\
  & \hspace{-7pt} \scriptsize{Number of Episodes}
    \end{tabular}}
  
    \caption{\footnotesize\textit{Regret analysis comparing the performance of KSH-LSPI models, where the candidate feature $x$ is independently represented using state features $\{\mathbf{s}_1, \mathbf{s}_2, \mathbf{s}_3\}$, and neural network-based approaches as described in Section \ref{neural-simulation-section}. Within each sub-figure, the dimensionality of the state space and the number of episodes used to generate the batch dataset are varied.}}
    \label{fig:neural_approaches_comparison}
  \end{figure}

\subsection{Comparison to Neural Approaches}
\label{neural-simulation-section}
We evaluate the performance of the KSH-LSPI algorithm against several widely-used neural network-based approaches, specifically:  neural fitted Q-iteration (NFQ), deep Q-network (DQN), double deep Q-network (DDQN), and conservative Q-learning (CQL) \citep{Mnih2013PlayingLearning, VanHasselt2010DoubleQ-learning, Riedmiller2005NeuralMethod, Kumar2020b}. Each model is trained using a batch dataset of experiences, which is gathered from a random policy interacting in an MDP with correlated state features and an additive reward function. Similar to Equation \eqref{additive-reward}, the reward function is dependent on the first two state features $\{\mathbf{s}_1, \mathbf{s}_2\}$ and the selected action $a$. Appendix A.1 provides a detailed description of the MDP and the data generation process (Emedom-Nnamdi, 2024) \nocite{AOAS-KSH-LSPI-Supp}. In each experiment, we adjust the dimensionality of the MDP's state space and the number of episodes used to generate the batch dataset. For the KSH-LSPI algorithm, we fit a separate model, where the candidate feature $x$ is represented as one of the first three state features $\{\mathbf{s}_1, \mathbf{s}_2, \mathbf{s}_3\}$. Here, each state feature, denoted as $\mathbf{s}_i$, contributes to the marginal component, $g_a(\mathbf{s}_i)$, as illustrated in Equation \eqref{simulation-additive-model}. We perform policy iteration in accordance with Example \ref{policy_iteration_discrete_action}, setting the maximum number of allowed policy iterations to $3$. Detailed specifications and architectures of both the KSH-LSPI models and the neural network-based approaches can be found in Appendix A.2 (Emedom-Nnamdi, 2024) \nocite{AOAS-KSH-LSPI-Supp}.

The estimated policies for each approach were evaluated within the MDP used to generate the training batch data set. Specifically, each policy was rolled out for 10 time steps (i.e., an episode), 1000 times. We performed a regret analysis, where, at the end of each episode, the difference between the optimal reward at each time step and the reward obtained by the current policy was calculated. The average of these differences over all episodes was then computed to obtain the estimated mean regret for each experiment. Figure \ref{fig:neural_approaches_comparison} presents results from the regret analysis, where the dimensionality of the state space and the number of episodes used to generate the batch dataset were varied. Within each experiment, we observe that the KSH-LSPI model under a candidate feature of $\mathbf{s}_2$ and $\mathbf{s}_3$, performs similarly to neural network-based approaches when the number of episodes is 100, and worse when the number of episodes used in the generated batch data set increases to 1000. Conversely, when $\mathbf{s}_1$ (i.e., the feature that accounts for the most variation within the observed rewards) is set as the candidate feature, the KSH-LSPI model outperforms the neural-network based models, and is further improved as the number of episodes increases. These results highlight a key sensitivity in the KSH-LSPI model; that is, the appropriate selection of the candidate feature $x$ largely influences model performance.

Lastly, the marginal performance differences among the neural network approaches may be attributable to the simulation environment. For example, in more complex, high-dimensional environments with offline data, we would expect approaches such as CQL to provide more significant benefits.
\section{Motivating Case Study}
\label{Motivating Case Study} 
Postoperative recovery is defined as the period of functional improvement that occurs from the end of surgery and hospital discharge to the instance in which normal function has been restored \citep{Bowyer2016PostoperativeWhom}. Depending on the type of surgery administered, this period of functional recovery can vary drastically and be accompanied by mild to severe complications. For patients who received corrective surgery for spine disease, postoperative recovery is impacted by the complexity of the diagnosis and surgical procedure received. Additional barriers to recovery for spine disease patients include stress, pain, cognitive dysfunction, and potential postoperative complications \citep{Wainwright2016EnhancedSurgery}. 

To improve postoperative recovery and care for spine patients, physicians employ a multipronged approach focusing on protocols that accelerate functional recovery, decrease postoperative complications, and improve subjective patient experience \citep{Elsarrag2019EnhancedReview}. As part of this effort, patient mobilization and consistent pain management are heavily suggested \citep{Burgess2019WhatReview}. To advance these efforts, physicians require objective measurements of a patient's functional capacity and pain over the course of their recovery \citep{Cote2019, Panda2020SmartphoneSurgery, Karas2020PredictingWearables, Boaro2021SmartphoneProject}. With respect to spine patients, such measurements can provide a formal understanding and quantification of mobilization activities that expedite overall patient recovery and minimize the risk of complications.

\begin{table}
    \caption{\label{table-stats} Subset of GPS and accelerometer-based summary statistics. Definitions can be found on the \textit{Forest} GitHub repository (\url{www.github.com/onnela-lab/forest}).}
    \centering
    \scalebox{0.85}{
    \begin{tabular}{||ccc||}\toprule  
        Distance Traveled (km) & Radius of Gyration (km) & Average flight duration (km)\\\midrule
        Time Spent at Home (hours) & Maximum Diameter (km)  & Fraction of the day spent stationary\\  \midrule
        Max. Distance from Home (km) &  Num. Significant Places Visited & Time Spent Walking \\  \midrule
        Average flight length (km) & Number of Steps  & Average Cadence\\   \bottomrule
    \end{tabular}}
    \label{tab:my_label}
\end{table}

 \begin{figure}[t]
    \centering
    \includegraphics[width=0.95\linewidth]{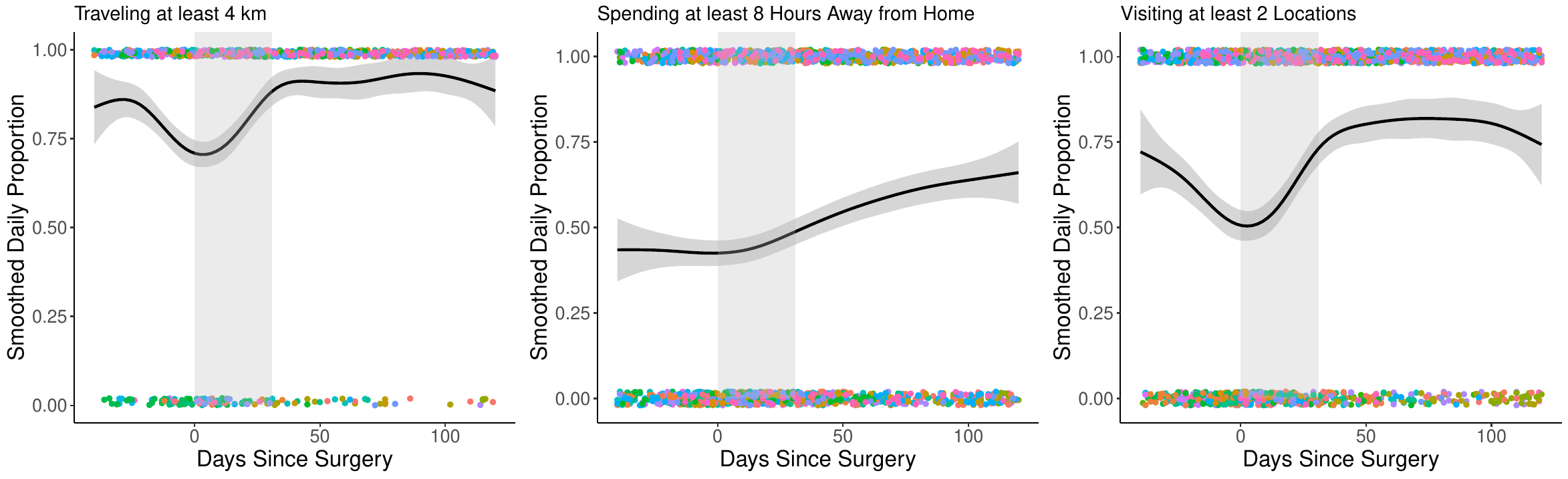} \\
    \caption{\label{fig:smooth}\footnotesize\textit{Smoothed mobility proportions (with standard errors represented in grey) for spine disease cohort centered on day of surgery. The lighter shaded area corresponds to the first 30 postoperative days. Receiving a neurosurgical intervention is followed by a period of decreased mobility, where patients tend to travel less and stay at home over a longer duration. Individual-level differences in recovery are driven by factors such as the type of surgery received, the specific diagnosis of spine disease, and patient demographics.}}
  \end{figure} 
 
We consider $n=67$ neurosurgical spine patients with a median age of 57 years (IQR: 48--65.5) who were enrolled between June 2016 and March 2020 as part of a mobile health study at Brigham and Women's Hospital. Each patient underwent a neurosurgical intervention in relation to their spine disease. For data collection, patients installed the Beiwe application on their smartphones. Beiwe is a high-throughput research platform that was developed by the Onnela lab at Harvard T.H. Chan School of Public Health for smartphone-based digital phenotyping on iOS and Android devices. Passive features collected on Beiwe include GPS and accelerometer data in their raw unprocessed form, Bluetooth and WiFi logs, and anonymized phone call and text message logs. Samples were collected from the GPS data stream for 1 minute every 5 minutes and from the accelerometer data stream for 10 seconds every 10 seconds. Using raw data sampled from GPS and accelerometer sensors, a set of behavioral characteristics is calculated regarding patient mobility at the daily level \citep{Liu2021BidirectionalProcess}. A subset of these features are represented in Table \ref{table-stats}. For active data collection, patients were electronically surveyed once daily at 5PM Eastern standard time to evaluate their current pain level. The prompt of the micro-survey was "Please rate your pain over the last 24 hours on a scale from 0 to 10, where 0 is no pain at all and 10 is the worst pain imaginable."
  
\begin{figure}[t]
    \centering
    \includegraphics[width=\linewidth]{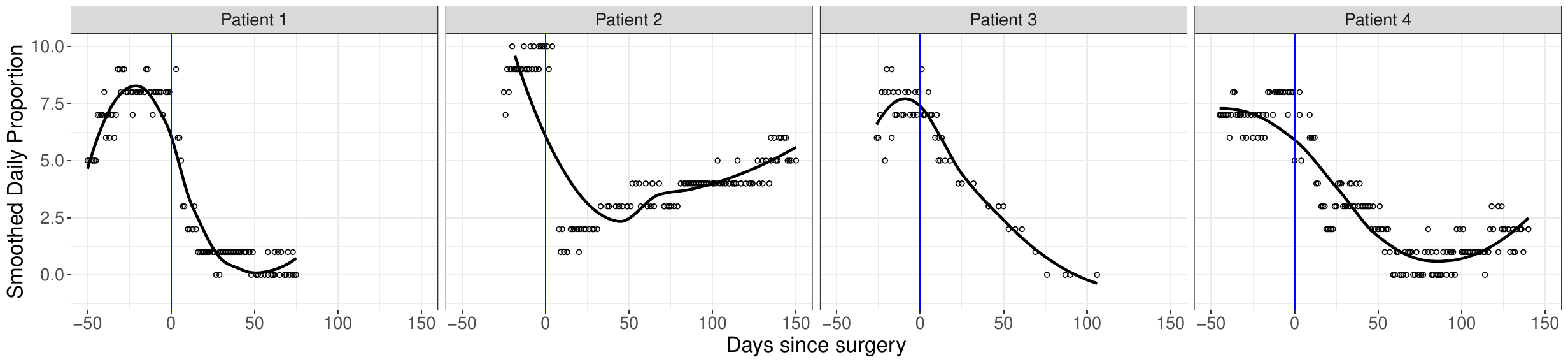} \\
    \caption{\label{fig:scatter}\footnotesize\textit{Pre- and postoperative pain responses with time centered on the day of surgery (i.e., blue line) with a fitted local regression (i.e., black line) for a random selection of patients. While surgery corresponds to a sharp decline in self-reported pain, we observe a heterogeneous recovery experience among these four patients.}}
\end{figure}
 
\indent In conjunction with the daily self-reported micro-surveys, these constructed features allow researchers to objectively identify postoperative trends in mobility and pain as it relates to overall functional recovery \citep{Boaro2021SmartphoneProject, Cote2019}. To this end, we leverage reinforcement learning to estimate and interpret mobility-based action-value functions that provide recommendations concerning questions such as "What level of mobilization is advisable after surgery?" and "How should these levels be adjusted given a patient's current condition?". The overall goal of these recommendations is to manage a patient's overall pain level and promote improved recovery. Furthermore, by utilizing an interpretable representation of the estimated action-value function, we seek to identify clinical and behavioral features that are important to consider for decision-making.

\section{Application to Surgical Recovery}
\label{application}

Using data from the spine disease cohort in Section \ref{Motivating Case Study}, we use nonparametric additive models to estimate action-value functions associated with 
\vspace{-0.25em}
\begin{enumerate}
    \item\label{A1} A \textbf{behavioral policy} that aims to mimic decisions commonly taken by patients, and 
    \item An \textbf{improved policy} retrieved from performing approximate policy iteration on the estimated behavioral policy.
\end{enumerate}
\vspace{-0.25em}
In both cases, the estimated decision-making policy aims to suggest the daily number of steps necessary to reduce long-term ($\gamma \gg 0$) postoperative pain response. We explore both discrete and continuous action spaces and provide a practical interpretation of the additive functional components as presented in Equations \eqref{eq-2.1} and \eqref{eq-2.2}, respectively.

\subsection{Data Pre-processing}
 We consider the recovery period of $n=67$ neurosurgical spine disease patients with a mean postoperative follow-up of 87 days (SD = 51.21 days). Baseline clinical information on this study cohort can be found in Table \ref{dp-demographic}. Behavioral features, derived from raw GPS and accelerometer data, were summarized on a daily time scale to closely monitor each patients' clinical recovery and/or progression after surgery. These features include passively sampled summary statistics that uniquely describe a patient's daily mobility and activity levels. 

We construct a simple MDP where each time step $t$ corresponds to a day since surgery. The state space, $S \in \mathbb{R}^d$ is a multidimensional, continuous state vector that consists of relevant behavioral features and patient-specific demographic information (i.e., age and days since surgery). In total, $d=9$ features were used in this analysis.\footnote{These features include: age, number of days since surgery, time spent at home (hours), distance traveled (km), maximum distance from home (km), radius of gyration (km), average cadence, and time spent not moving (hours).} The action space $\mathcal{A} \in \mathbb{R}$ represents the number of steps taken per day. For the discrete action model, the action space, $\mathcal{A} \in \{0,1\}$, is binarized such that $0$ represents moving less than the subject-level preoperative median number of steps taken per day and $1$ represents moving above this threshold. The rewards, $r \in \mathbb{R}$, are chosen to be the negative value of the self-reported pain score, where each score is taken from a numerical rating scale between 0 (i.e., no pain) and 10 (i.e., worst pain imaginable). Lastly, we consider a discount factor $\gamma$ of 0.5 to examine estimated policies that aim to reduce long-term pain response. 

\begin{table}[t]
      \caption{Participant demographic information and mobile health data for spine disease cohort. Summaries are computed according to the first 60 postoperative days since surgery (including the day of surgery).}
\centering 
  \begin{threeparttable}
  \footnotesize
    \begin{tabular}{m{60mm} m{45mm}}
    Variable  & $n$ $(\%)$ or Median ($25^\text{th}$--$75^\text{th}$)  \\
     \midrule\midrule
       \textbf{Demographic Data} &  \\[0.25em]
        \hspace{1em}Age &     57.0 (48.0--65.5)   \\
        \hspace{1em}Female gender & 34 (50.7) \\ 
    \cmidrule(l r ){1-2}
     \textbf{Site of surgery} &   \\[0.25em] 
     \hspace{1em}Cervical &  19 (28.4) \\ 
     \hspace{1em}Lumbar & 27 (40.3) \\ 
     \hspace{1em}Thoracic & 2 (3.0)   \\
     \hspace{1em}Multiple & 18 (26.9)\\
    \cmidrule(l r ){1-2}
    \textbf{Data Collection} &  \\[0.25em] 
    \hspace{1em}GPS days of follow-up &  61 (49--61)      \\
    \hspace{1em}Accelerometer days of follow-up  &  61 (50.5--61)      \\
    \hspace{1em}Daily pain survey response rate & 59.4 (42.4--76.9) \\ 

    \cmidrule(l r ){1-2}
    \textbf{Digital Phenotypes} &  \\[0.25em] 
        \hspace{1em}Number of places visited & 3 (2--5) \\ 
    \hspace{1em}Time spent at home (hours) &  18.3 (12.9--21.9)\\ 
     \hspace{1em}Distance traveled (km) & 32.3 (10.8--62.3) \\ 
     \hspace{1em}Maximum distance from home (km)&  10.6 (4.5--25.5) \\ 
     \hspace{1em}Radius of gyration (km)&  1.50 (0.18-5.01) \\ 
     \hspace{1em}Time spent not moving &  21.2 (20.2--22.2) \\ 
    \hspace{1em}Average cadence & 1.64 (1.55--1.74) \\ 
     \hspace{1em}Number of steps & 948.6 (356.9--2,005) \\ 
    \midrule\midrule
    \end{tabular}
\end{threeparttable}
      \label{dp-demographic}
\end{table}

Under this MDP, we consider up to the first 60 days since surgery for each patient. Patients with a postoperative follow-up period of less than 5 days were excluded. Entries with missing values in either the behavioral features or the daily self-reported pain scores were removed. The batch dataset $\mathcal{D}$ with $N=$ 1,409 daily transitions was constructed using data collected from the study cohort and represented using the MDP. All state features were normalized to a [0,1] range for model fitting. 

\subsection{Model Fitting}

To estimate the action-value function associated with the behavioral policy $\pi_b$, we use Algorithm \ref{KSH-LSTD} and construct $\boldsymbol{\Phi}'$ using the observed next-state action contained within each patient-level trajectory in $\mathcal{D}$. That is, $\boldsymbol{\Phi}_{i\bigcdot}' = \phi(\mathbf{s}^{'[i]}, a^{[i+1]})$ for the $i^{\text{th}}$ observed transition. Accordingly, we estimate the action-value function for the improved policy $\pi^*$ using approximate policy iteration (as detailed Algorithm \ref{KSH-LSPI}) on the action-value function associated with the behavioral policy. For both discrete and continuous action versions of the general model \eqref{general-model}, we use a Gaussian kernel $K(u)= e^{-\frac{1}{2} u^{2}}$ and a grid of evenly spaced points $\mathcal{Z}$ within a [0,1] range for discretization. In the discrete action model, we estimate the marginal effect $g_a(x)$ and the additive joint effects $f_{j,a}(\mathbf{s}_j,x)$ for $j\geq 2$ for each candidate state feature $x$ in a set $\mathcal{H}$, where $x \neq \mathbf{s}_j$.

To select the hyperparameters of the KSH-LSTDQ estimator (i.e., the degree of the B-spline functions, the number of basis functions, the bandwidth, and the regularization penalty), we partitioned the dataset $\mathcal{D}$ into training and validation sets according to a $80\%-20\%$ patient-level split and performed a grid search. Using these partitions, we retrieved a set of hyperparameters that minimized the validation mean squared error between the estimated action-value under the behavioral policy when $\gamma = 0$ and the true immediate rewards,
\begin{align*}
    \text{MSE}(\mathcal{D}_\text{Val}) = \frac{1}{|\mathcal{D}_\text{Val}|} \sum_{(\mathbf{s},a,r)\sim \mathcal{D}_\text{Val}} \Big( \hat{Q}^{\pi_b}(\mathbf{s},a) - r\Big)^2 .
\end{align*} 
Accordingly, these hyperparameters were used to retrieve the KSH-LSTDQ estimators for MDPs where $\gamma$ is set to 0.5. The set of hyperparameters used for each estimated model is shown in Appendix A.2 (Emedom-Nnamdi, 2024) \nocite{AOAS-KSH-LSPI-Supp}. 


\subsection{Results and interpretations}
We visualize and interpret the estimated additive component functions of the action-value functions associated with the behavioral and improved policies.
\subsubsection{Discrete Action Model}
In the discrete action model, we estimate a nonparametric action-value function for each candidate state feature in the set $\mathcal{H}$, which we represent as $x$. Here, $\mathcal{H}$ consists of the following features: age, number of days since surgery, time spent at home (hours), and distance traveled (km). 

\begin{figure}[bt!]
\centering
\begin{center}
    {\tiny Move \textbf{ABOVE} preoperative baseline ($a=1$):} \textcolor{Green}{\threeemdash} \hspace{0.75em} {\tiny Move \textbf{BELOW} preoperative baseline ($a=0$):} \textcolor{Purple}{\threeemdash}
\vspace{-1em}
\end{center}
\vspace{-0.5em}
\subfloat[\textbf{Behavioral policy}]{
\begin{tabular}{cccc}
  {\tiny\rotatebox{90}{\hspace{12pt}$\text{Marginal Effect}$}} &\hspace{-12pt}  \includegraphics[width=0.20\linewidth]{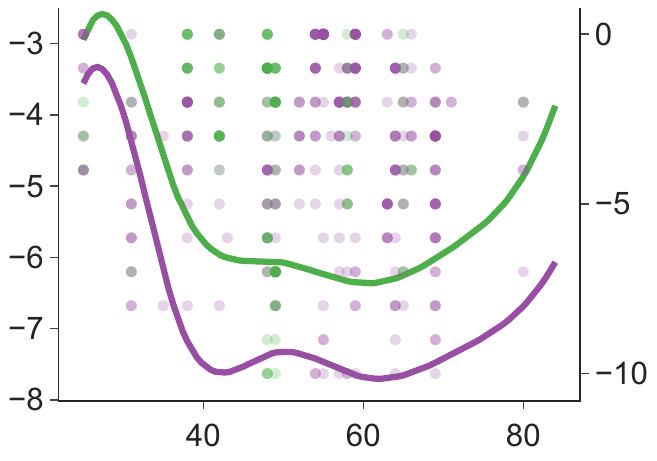} & \hspace{-10pt}
  \includegraphics[width=0.20\linewidth]{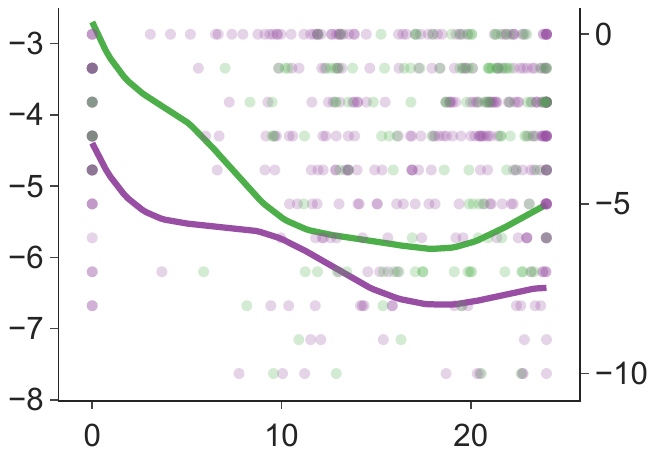} \vspace{-0.75em}& {\hspace{-25pt}\tiny\rotatebox{90}{\hspace{3pt}$\text{Negative Pain Score}$}}\\
  &\hspace{-10pt}\tiny$x = \text{Age}$ & 
  \hspace{-7pt}\tiny$x = \text{Time spent at home (hours)}$& \\
  {\tiny\rotatebox{90}{\hspace{12pt}$\text{Marginal Effect}$}} &\hspace{-12pt} \includegraphics[width=0.20\linewidth]{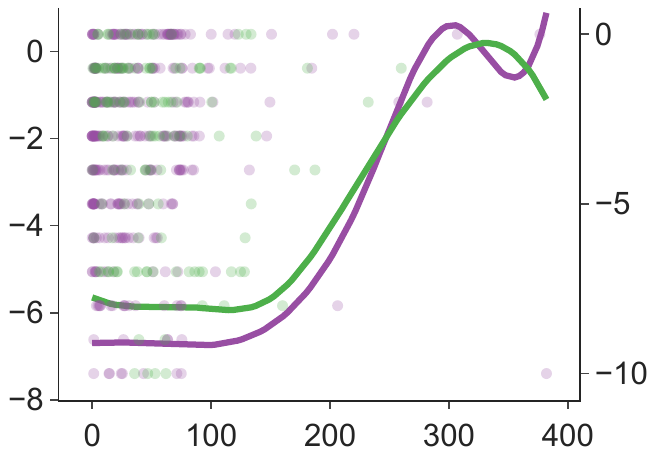} & \hspace{-10pt}
 \includegraphics[width=0.20\linewidth]{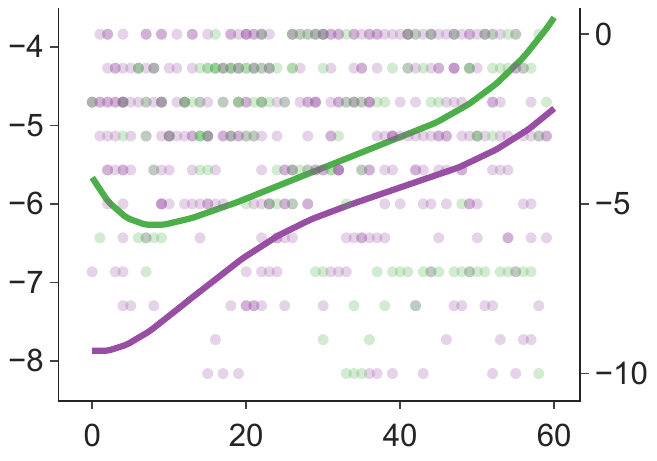} 
 \vspace{-0.75em}& {\hspace{-25pt}\tiny\rotatebox{90}{\hspace{3pt}$\text{Negative Pain Score}$}}\\
  &\hspace{-10pt}\tiny$x = \text{Distance traveled (km)}$ & 
  \hspace{-10pt}\tiny$x = \text{Days since surgery}$&
\end{tabular}\label{fig:marginal-effect-discrete-0.5-behavioral}
 }
\subfloat[\textbf{Policy Iteration}]{
\begin{tabular}{cccc}
  {\tiny\rotatebox{90}{\hspace{12pt}$\text{Marginal Effect}$}} &\hspace{-12pt} \includegraphics[width=0.20\linewidth]{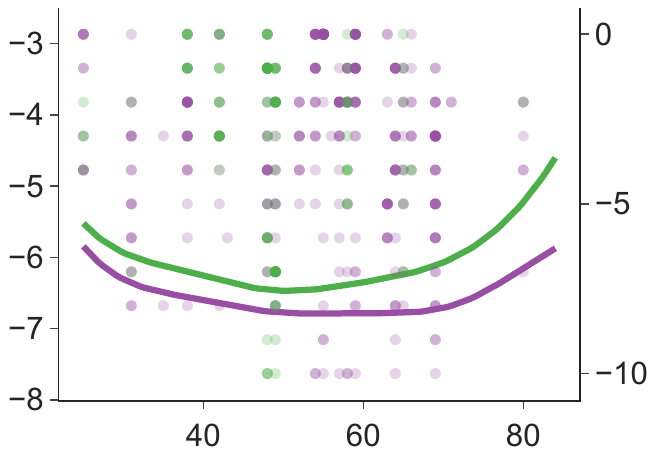} & \hspace{-10pt}
  \includegraphics[width=0.20\linewidth]{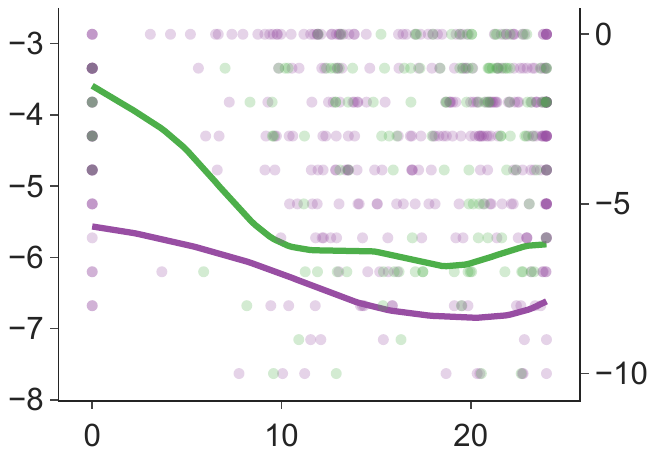} 
  \vspace{-0.75em}& {\hspace{-25pt}\tiny\rotatebox{90}{\hspace{3pt}$\text{Negative Pain Score}$}}\\
  &\hspace{-10pt}\tiny$x = \text{Age}$ & 
  \hspace{-7pt}\tiny$x = \text{Time spent at home (hours)}$& \\
  {\tiny\rotatebox{90}{\hspace{12pt}$\text{Marginal Effect}$}} &\hspace{-12pt} \includegraphics[width=0.20\linewidth]{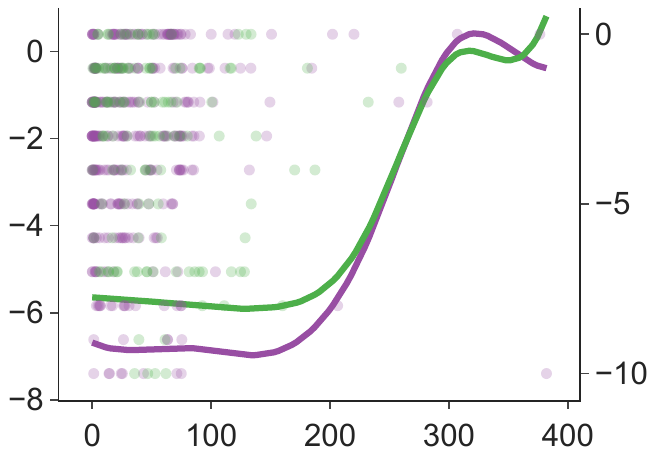} & \hspace{-10pt}
 \includegraphics[width=0.20\linewidth]{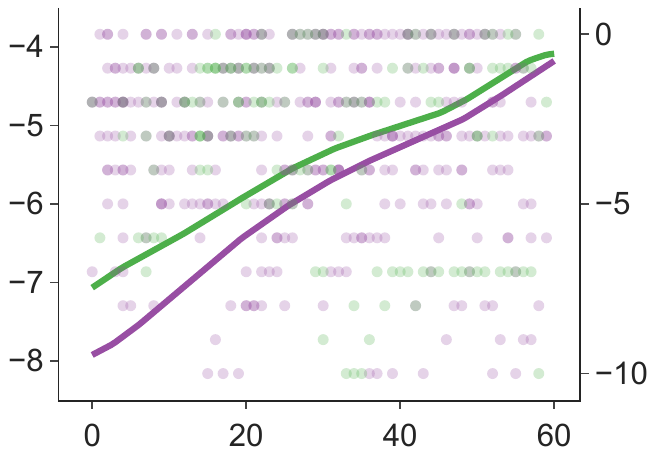}
  \vspace{-0.75em}& {\hspace{-25pt}\tiny\rotatebox{90}{\hspace{3pt}$\text{Negative Pain Score}$}}\\
  &\hspace{-10pt}\tiny$x = \text{Distance traveled (km)}$ & 
  \hspace{-10pt}\tiny$x = \text{Days since surgery}$&
\end{tabular}\label{fig:marginal-effect-discrete-0.5-improved}
}

\caption{\footnotesize\textit{A comparison of the marginal component function $\hat{g}_a(x)$ of $Q^\pi(\mathbf{s},a, x)$ estimated under the behavioral policy $\pi=\pi_b$ vs. the improved policy $\pi=\pi^*$. Each sub-figure is associated with a separate nonparametric additive model of $Q^\pi(\mathbf{s},a, x)$, where the state feature representing $x$ is changed. For each action, the solid lines represent the estimate of the marginal component function $\hat{g}_a(x)$ over the range of observed values of $x$, whereas the points represent the value of the associated observed rewards (i.e., negative pain score) over $x$.}}
\label{fig:marginal-effect-discrete-0.5}
\end{figure}

\textit{Marginal State Feature Effects.} In Figure \ref{fig:marginal-effect-discrete-0.5}, we examine the marginal effect $\hat{g}_a(x)$ of the estimated action-value function for each candidate state feature $x \in \mathcal{H}$ under the behavioral policy $\pi_b$ and the improved policy $\pi^*$ constructed using policy iteration. Specifically, $\hat{g}_a(x)$  estimates the marginal change in \textit{long-term} negative pain response for each state feature $x$ under action $a$. Across each sub-figure in Figure \ref{fig:marginal-effect-discrete-0.5-behavioral}, moving above a patient's preoperative baseline number of steps ($a=1$) within a given day is associated with a higher immediate negative pain response in comparison to the converse action ($a=0$), regardless of the value of $x$. This observation is in line with clinical research that suggests movement at or above a patient's preoperative baseline is associated with improved postoperative functional recovery \citep{ Duc2013ObjectiveActivity., Ozkara2015EffectivenessSurgery., Cote2019}.

\begin{figure}[bt!]
    \centering
    \hspace{-1em}
\subfloat[\textbf{Behavioral policy}]{\includegraphics[width=0.35\linewidth]{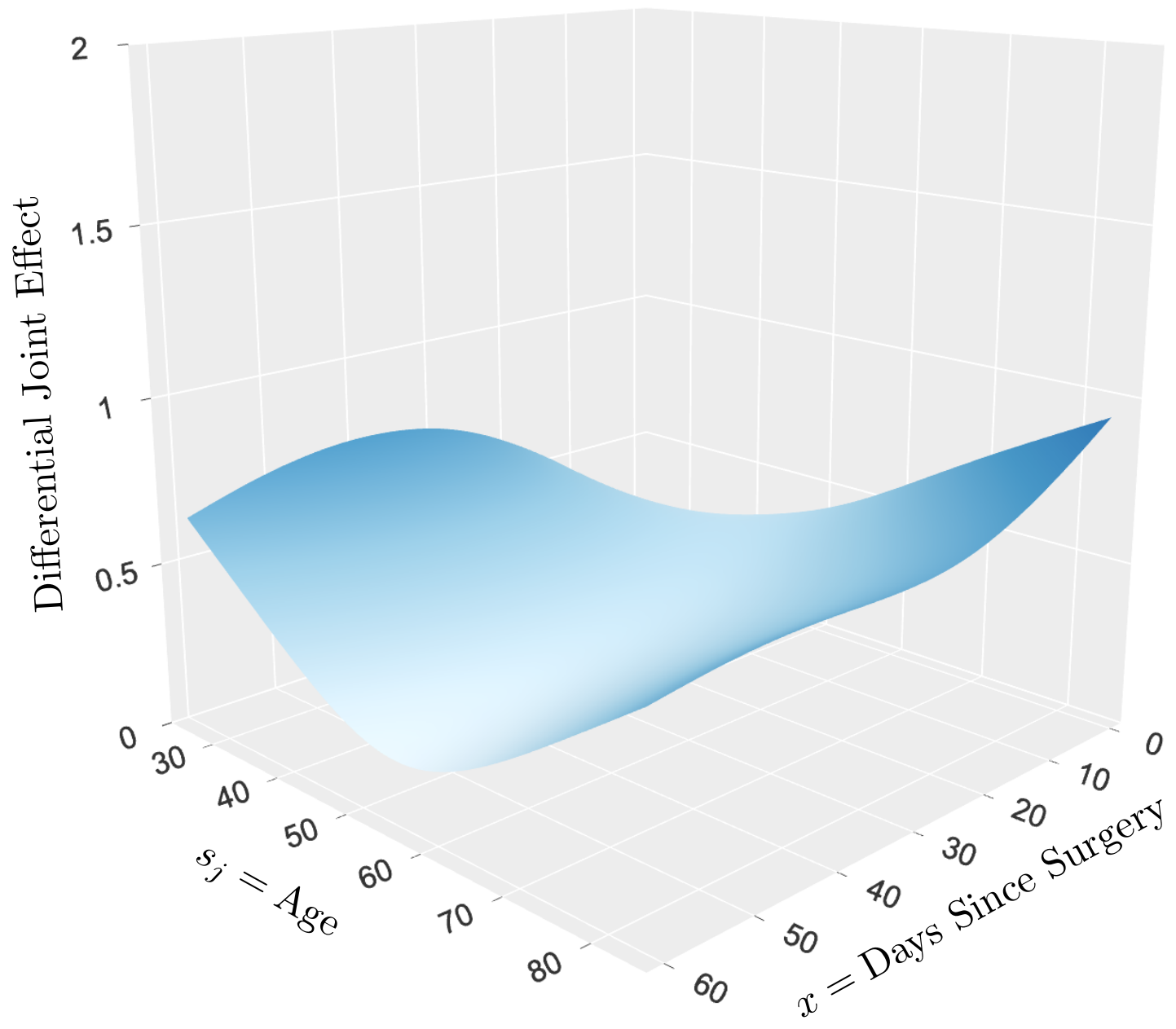}  }
\hspace{4em}
\subfloat[\textbf{Policy Iteration}]{\includegraphics[width=0.35\linewidth]{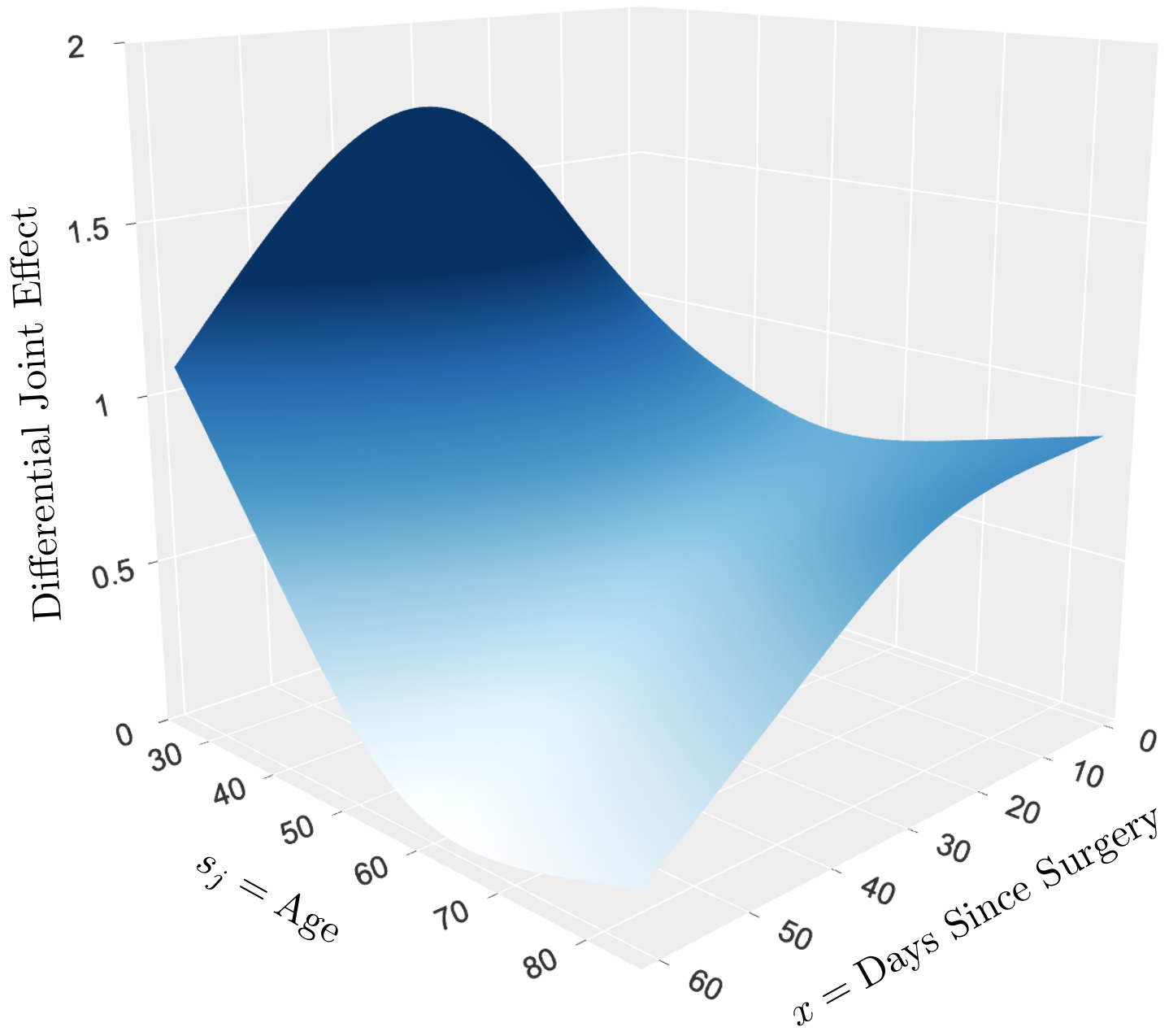}}

    \caption{\footnotesize\textit{Surface plots representing the difference between the joint component functions $\hat{f}_{j,1}(\mathbf{s}_j,x)$ and $\hat{f}_{j,0}(\mathbf{s}_j,x)$ (i.e., the differential benefit of selecting action $a$=1 over $a=0$) of $Q^\pi(\mathbf{s},a,x)$ estimated under the behavioral policy $\pi=\pi_b$ vs. the improved policy $\pi=\pi^*$. }}
\label{fig:joint-effect-discrete-0.5-surface}

\centering
\subfloat[\textbf{Behavioral policy}]{
\begin{tabular}{l c c c}
     \raisebox{\height}{{\tiny\rotatebox{90}{\hspace{13pt}$\text{Age}$}}}  & \hspace{-13pt} \includegraphics[width=0.195\linewidth]{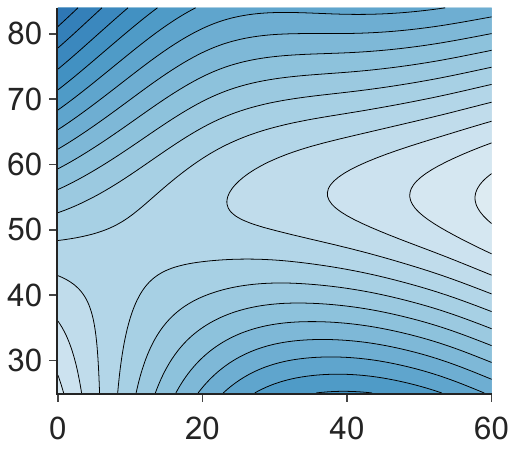} & \hspace{-0.75em}{\tiny\rotatebox{90}{\hspace{5pt}$\text{Time spent not moving}$}} & \hspace{-10pt}\includegraphics[width=0.19\linewidth]{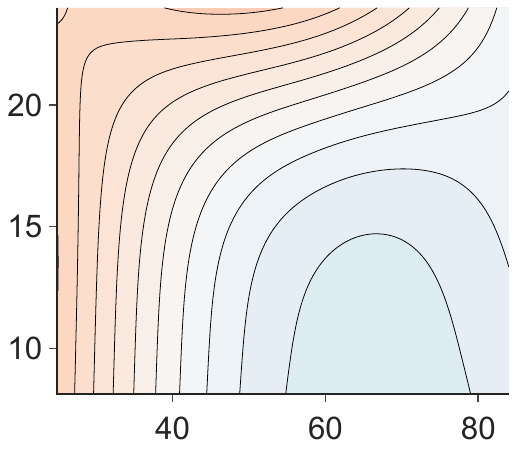} \vspace{-0.75em}\\ 
     \multicolumn{2}{c}{\hspace{15pt}\tiny$x = \text{Days since surgery}$} & \multicolumn{2}{c}{\hspace{-3pt}\tiny$x = \text{Age}$} \vspace{0.2em}\\
  {\tiny\rotatebox{90}{\hspace{16pt}$\text{Average Cadence}$}} & \hspace{-13pt}\includegraphics[width=0.19\linewidth]{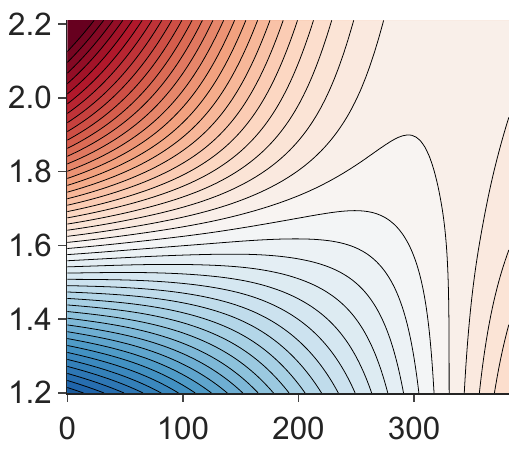} &\hspace{-0.75em}{\tiny\rotatebox{90}{\hspace{7pt}$\text{Max. dist. from home}$}} &\hspace{-10pt}\includegraphics[width=0.19\linewidth]{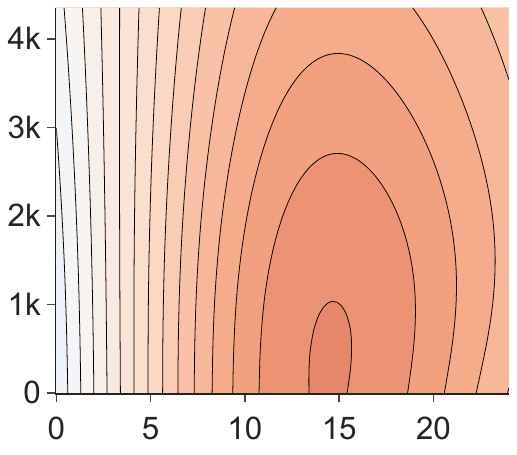}
\vspace{-0.75em}\\
     \multicolumn{2}{c}{\hspace{15pt}\tiny$x = \text{Distance travel}$} & \multicolumn{2}{c}{\hspace{-5pt}\tiny$x = \text{Time spent at home}$}
\end{tabular}}
\subfloat[\textbf{Policy Iteration}]{
\begin{tabular}{l c c c c}
     \raisebox{\height}{{\tiny\rotatebox{90}{\hspace{13pt}$\text{Age}$}}}  & \hspace{-13pt} \includegraphics[width=0.195\linewidth]{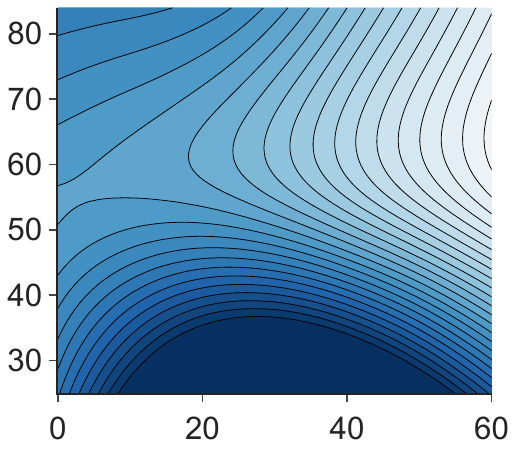} & \hspace{-0.75em}{\tiny\rotatebox{90}{\hspace{5pt}$\text{Time spent not moving}$}} & \hspace{-10pt}\includegraphics[width=0.19\linewidth]{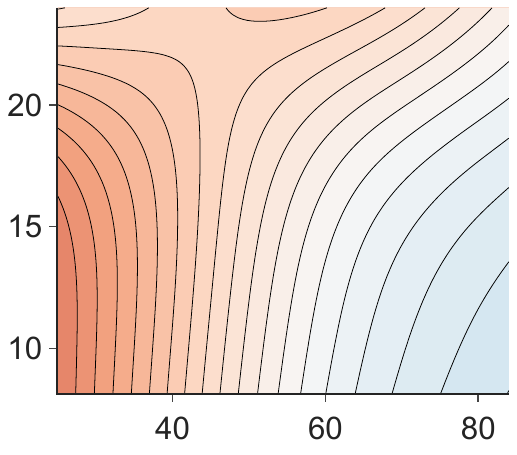} \vspace{-0.75em}\\
     \multicolumn{2}{c}{\hspace{15pt}\tiny$x = \text{Days since surgery}$} & \multicolumn{2}{c}{\hspace{-3pt}\tiny$x = \text{Age}$} \vspace{0.2em}\\
  {\tiny\rotatebox{90}{\hspace{16pt}$\text{Average Cadence}$}} & \hspace{-13pt}\includegraphics[width=0.19\linewidth]{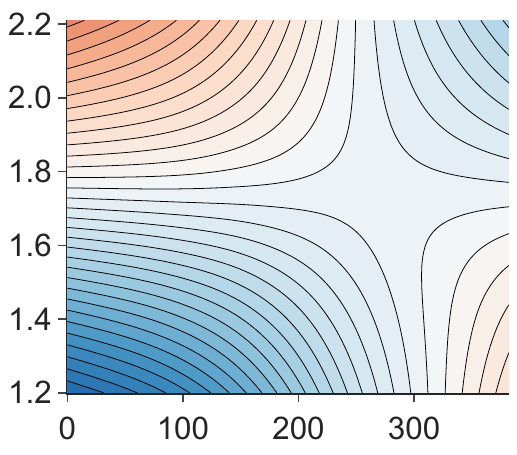} &\hspace{-0.75em}{\tiny\rotatebox{90}{\hspace{7pt}$\text{Max. dist. from home}$}} &\hspace{-10pt}\includegraphics[width=0.19\linewidth]{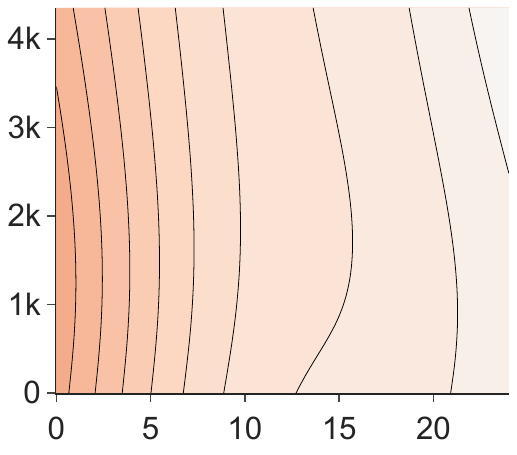}
\vspace{-0.75em}\\
     \multicolumn{2}{c}{\hspace{15pt}\tiny$x = \text{Distance travel}$} & \multicolumn{2}{c}{\hspace{-5pt}\tiny$x = \text{Time spent at home}$} & \multirow{4}{*}[1.85in]{\includegraphics[width=0.045\linewidth]{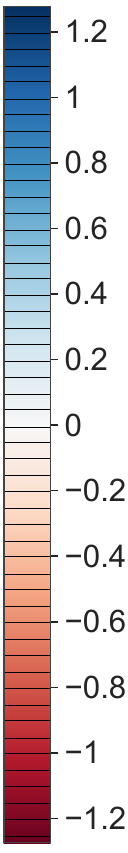}}
\end{tabular}}

\caption{\footnotesize\textit{Contour plots representing the differential benefit of selecting action $a=1$ over $a=0$ with respect to joint effects $\hat{f}_{j,a}( \mathbf{s}_j,x)$ under $Q^\pi(\mathbf{s},a,x)$ estimated under the behavioral policy $\pi=\pi_b$ vs. the improved policy $\pi=\pi^*$. Each sub-figure is associated with a separate nonparametric additive model of $Q^\pi(\mathbf{s},a,x)$, where the state feature representing $x$ is changed.}}
\label{fig:joint-effect-discrete-0.5}
\end{figure}

Within each select action, the marginal effect shows a nonlinear change in long-term pain response. When $x = \text{\textit{age}}$,  Figure \ref{fig:marginal-effect-discrete-0.5-behavioral} suggests a marginal increase in long-term negative pain response for younger and older ages. This observation supports clinical studies suggesting the existence of age-related pain sensitivity that peaks during mid-life \citep{Yezierski2012TheStudies.}. Additionally, when $x = \text{\textit{time spent at home}}$, our model suggests that spending more time at home is associated with a nonlinear decrease in negative pain response. Furthermore, when $x = \text{\textit{distance traveled}}$, we observe that, regardless of the selected action, traveling less than $150$km is associated with a constant effect on negative pain response, whereas traveling beyond $150$km within a given day is associated with an increasing effect. We note that this association is possibly due to survivorship bias present in the model estimates, where a few patients report minimal pain during periods of excessive travel.

Lastly, when $x = \text{\textit{days since surgery}}$, our model examines the impact of mobilization as the number of days since surgery increases. We note the difference in the marginal effect among both actions is maximized for days closest to the onset of surgery, suggesting that increased mobilization during early periods of recovery may be associated with decreased pain response. This finding supports current clinical practice that suggests early mobilization enhances surgical recovery, a cornerstone of postoperative pain management \citep{Wainwright2016EnhancedSurgery, Burgess2019WhatReview}. 

\newcommand{\shiftleft}[2]{\makebox[0pt][r]{\makebox[#1][l]{#2}}}
\newcommand{\shiftright}[2]{\makebox[#1][r]{\makebox[0pt][l]{#2}}}
\fboxrule=0.75pt

The differences between the marginal effects associated with the behavioral and improved policies, as shown in Figure \ref{fig:marginal-effect-discrete-0.5-improved}, are subtle. While the underlying trends and ordering of actions are relatively consistent, the estimated effect sizes appear to be smaller for select candidate state features under the improved policy (e.g., $x=$ age, time spent at home, or days since surgery) compared to those of the behavioral policy. 

\textit{Joint Effects betwen State Features.} In Figures \ref{fig:joint-effect-discrete-0.5-surface}  and \ref{fig:joint-effect-discrete-0.5}, we examine the joint effect $f_{j,a}(x, \mathbf{s}_j)$ of the estimated action-value function between select state features $\mathbf{s}_j$ (i.e., age, time spent not moving, average cadence, and maximum distance from home) and candidate state features $x \in \mathcal{H}$. The value of each joint feature pair corresponds to a nonlinear effect on an estimated smooth surface representing the additive, long-term change in negative pain response. We specifically examine the benefit of selecting a given action over its converse by visualizing the difference between the joint effects under both actions, i.e., $f_{j,1} - f_{j,0}$. Differences greater than zero indicate an additive preference for action $a=1$, over the converse $a=0$.

\begin{figure}[!t]
\centering

\subfloat[\textbf{Behavioral policy}]{
\begin{tabular}{ccc}
    {\scriptsize\rotatebox{90}{\hspace{33pt}$\text{Marginal Effect}$}} \hspace{-17pt} & \includegraphics[width=0.35\linewidth]{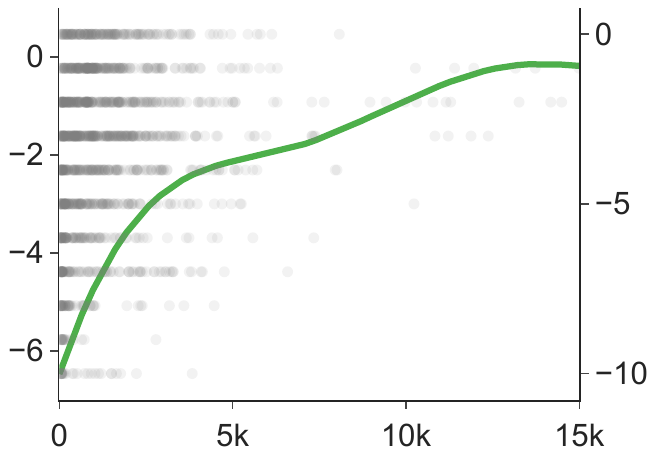} & {\hspace{-20pt}\scriptsize\rotatebox{90}{\hspace{28pt}$\text{Negative Pain Score}$}}  \\[-0.35em]
     & \scriptsize$a = \text{Step count}$ &
\end{tabular}  \label{fig:continous-marginal-effect-behavioral}
}
\hspace{0.1em}
\subfloat[\textbf{Policy Iteration}]{
\begin{tabular}{ccc}
    {\scriptsize\rotatebox{90}{\hspace{33pt}$\text{Marginal Effect}$}} \hspace{-17pt} & 
    \includegraphics[width=0.35\linewidth]{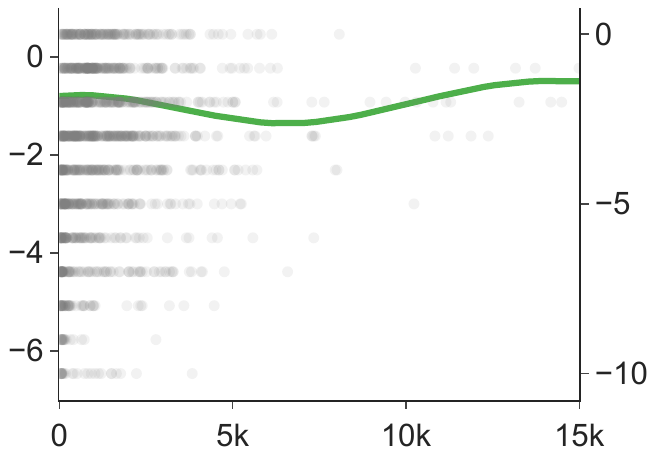} & {\hspace{-20pt}\scriptsize\rotatebox{90}{\hspace{28pt}$\text{Negative Pain Score}$}}  \\[-0.35em]
     & \scriptsize$a = \text{Step count}$ &
\end{tabular}  \label{fig:continous-marginal-effect-optimized}
}
\caption{\footnotesize\textit{The marginal component function $\hat{g}(a)$ of $Q^\pi(\mathbf{s},a)$ estimated under the behavioral policy $\pi=\pi_b$ vs. the improved policy $\pi=\pi^*$. The solid lines represent the estimate of the marginal component function $\hat{g}(a)$ over the observed values $a$, whereas the points represent the value of the associated observed reward (i.e., negative pain score).}}
 \label{fig:continous-marginal-effect}
\end{figure}

When examining the joint effect between $x = \text{\textit{days since surgery}}$ and $\mathbf{s}_j=\text{\textit{age}}$, we observe that regardless of the value of each corresponding feature, moving more than the preoperative baseline is associated with an increase in negative pain response throughout the domain of the joint component function. Interestingly, this association is more pronounced among younger patients under the improved policy. This observation is consistent with clinical research suggesting a relationship between increased postoperative movement and improved rehabilitation, and its potential modification by factors such as age \citep{Ozkara2015EffectivenessSurgery., Duc2013ObjectiveActivity., Jaensson2019FactorsReview}. 

When $x = \text{\textit{distance traveled}}$ and $\mathbf{s}_j=\text{\textit{average cadence}}$,  maintaining a slower average walking cadence over longer distances seems to be associated with the suggested action of moving beyond the preoperative baseline step count. This association is relatively consistent across both the behavioral and improved policies. However, under the improved policy, a positive association is also noticed with faster walking cadences near the upper boundary of total distance traveled. In general, the differential relationship between $x = \text{\textit{distance traveled}}$ and $\mathbf{s}_j=\text{\textit{average cadence}}$ could be indicative of the shift from automaticity to executive control of locomotion, as seen in rehabilitation literature \citep{Clark2015AutomaticityStrategies.}. This shift may occur as distances increase or as walking becomes more challenging (e.g., due to physical exertion, elevated pain response, or injury), requiring individuals to expend more cognitive effort (i.e., executive control) to manage their gait. 

\subsubsection{Continuous Action Model}
We estimate a nonparametric action-value function for continuous actions, specifically the number of steps taken, under both behavioral and improved policies.

\textit{Marginal effects.} In Figure \ref{fig:continous-marginal-effect}, we examine the marginal effect $\hat{g}(a)$ of the estimated action-value function. Specifically, $\hat{g}(a)$ estimates the marginal change in \textit{long-term} negative pain response for a select value of $a$, or number of steps taken. As with the discrete action model, we observe a positive association between the number of steps taken and the long-term negative pain response in the continuous action model, especially under the behavioral policy. For the behavioral policy (as shown in Figure \ref{fig:continous-marginal-effect-behavioral}), we observe that the marginal effect is log-shaped and increases with number of steps taken. For the improved policy (Figure \ref{fig:continous-marginal-effect-optimized}), the marginal effect remains relatively constant across the observed number of steps. 

\begin{figure}[!t]
 \centering
    \hspace{-1em}
\subfloat[\textbf{Behavioral policy}]{\includegraphics[width=0.35\linewidth]{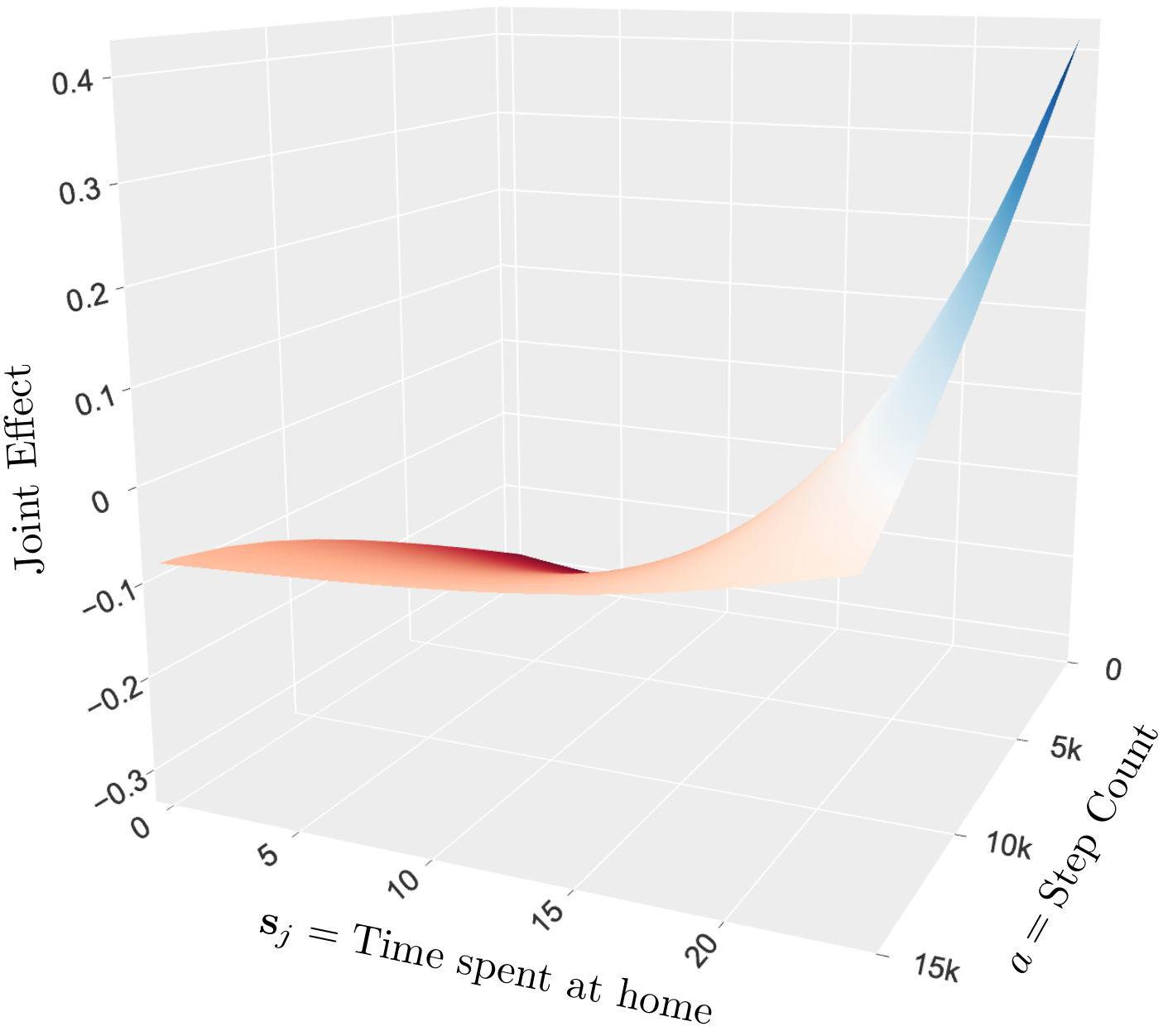}  \label{fig:joint-effect-cont-action-surface-behavioral-1}}
\hspace{4em}
\subfloat[\textbf{Policy Iteration}]{\includegraphics[width=0.35\linewidth]{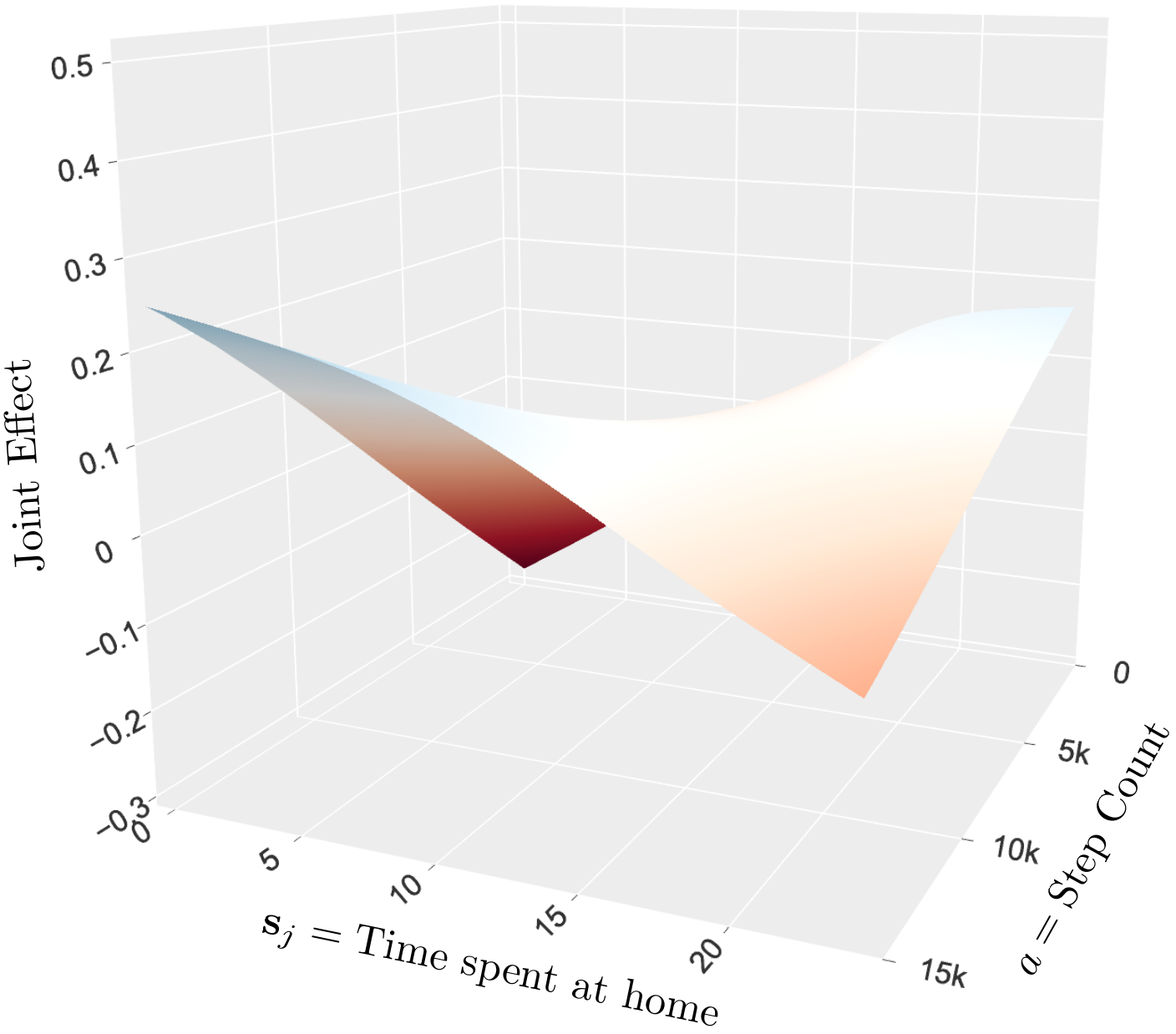} \label{fig:joint-effect-cont-action-surface-behavioral-2}}

    \caption{\footnotesize\textit{Surface plots representing the joint component functions $\hat{f}_{j}(\mathbf{s}_j,a)$ of $Q^\pi(\mathbf{s},a)$ estimated under the behavioral policy $\pi=\pi_b$ vs. the improved policy $\pi=\pi^*$. }}
\label{fig:joint-effect-cont-action-surface}

\subfloat[\textbf{Behavioral policy}]{
\begin{tabular}{l c c c}
     \raisebox{\height}{{\tiny\rotatebox{90}{\hspace{-18pt}$\text{Time spent at home}$}}}  & \hspace{-13pt} \includegraphics[width=0.19\linewidth]{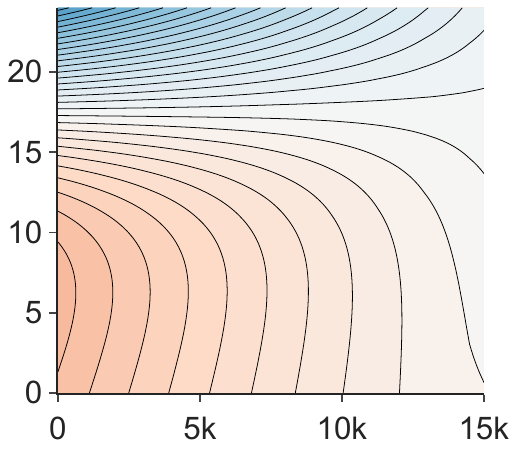} & \hspace{-1em}{\tiny\rotatebox{90}{\hspace{14.5pt}$\text{Distance traveled}$}} & \hspace{-15pt}\includegraphics[width=0.1965\linewidth]{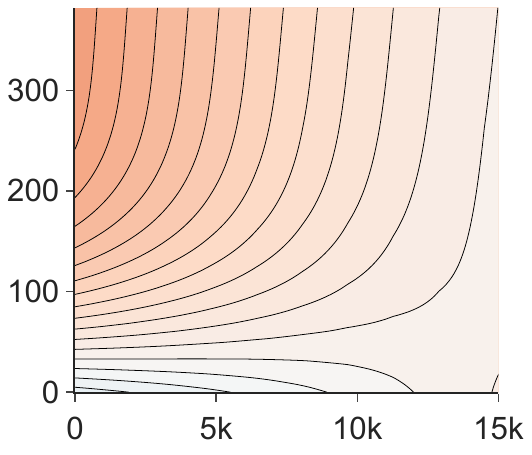} \vspace{-0.25em}\\
  {\tiny\rotatebox{90}{\hspace{35pt}$\text{Age}$}} & \hspace{-9pt}\includegraphics[width=0.1925\linewidth]{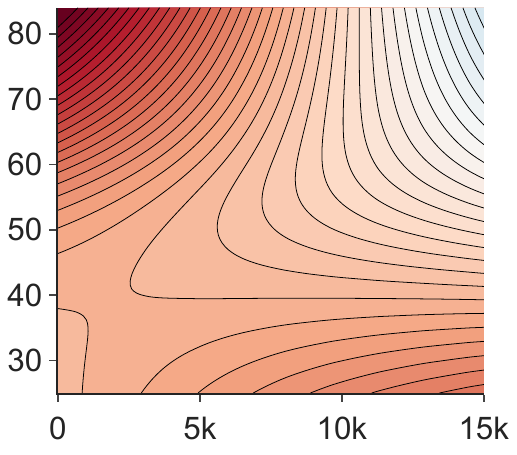} &\hspace{-0.85em}{\tiny\rotatebox{90}{\hspace{14pt}$\text{Days since surgery}$}} &\hspace{-10pt}\includegraphics[width=0.195\linewidth]{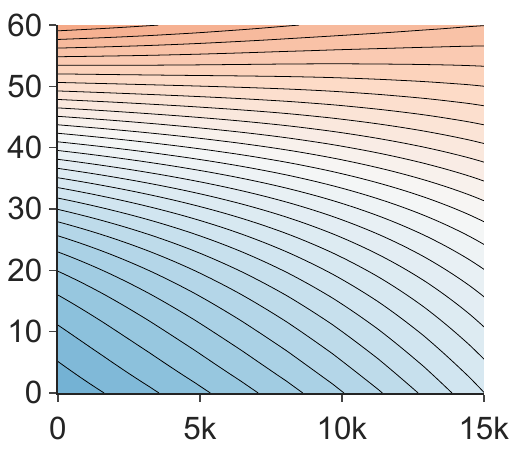}
\vspace{-0.75em}\\
     \multicolumn{2}{c}{\hspace{15pt}\tiny$a = \text{Step count}$} & \multicolumn{2}{c}{\hspace{-5pt}\tiny$a = \text{Step count}$}
\end{tabular} \label{fig:joint-effect-cont-action-behavioral}\hspace{-0.5em}}%
\subfloat[\textbf{Policy Iteration}]{
\begin{tabular}{l c c c c}
\raisebox{\height}{{\tiny\rotatebox{90}{\hspace{-18pt}$\text{Time spent at home}$}}}  & \hspace{-13pt} \includegraphics[width=0.19\linewidth]{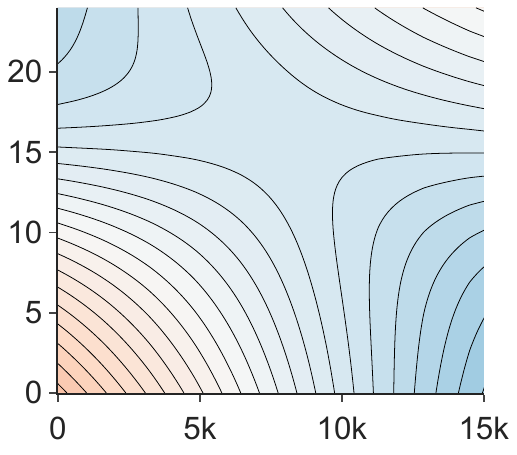} & \hspace{-1em}{\tiny\rotatebox{90}{\hspace{14.5pt}$\text{Distance traveled}$}} & \hspace{-15pt}\includegraphics[width=0.1975\linewidth]{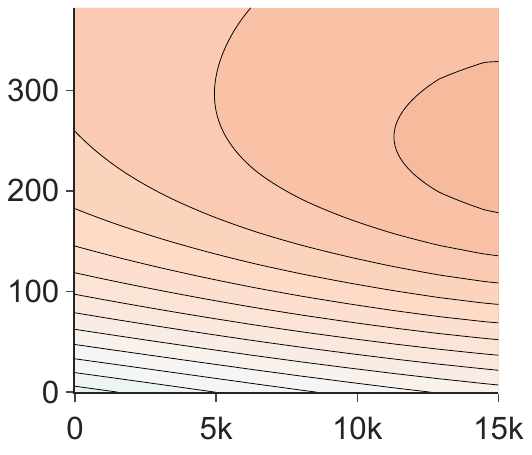} \vspace{-0.25em}\\
 {\tiny\rotatebox{90}{\hspace{35pt}$\text{Age}$}} & \hspace{-9pt}\includegraphics[width=0.1925\linewidth]{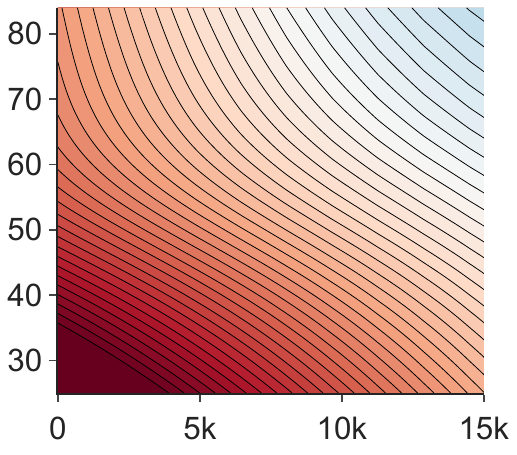} &\hspace{-0.85em}{\tiny\rotatebox{90}{\hspace{14pt}$\text{Days since surgery}$}} &\hspace{-10pt}\includegraphics[width=0.195\linewidth]{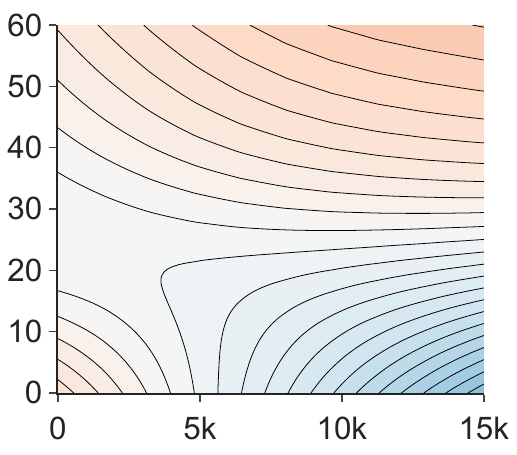}
\vspace{-0.75em}\\
     \multicolumn{2}{c}{\hspace{15pt}\tiny$a = \text{Step count}$} & \multicolumn{2}{c}{\hspace{-5pt}\tiny$a = \text{Step count}$} & \hspace{-5pt}\multirow{4}{*}[1.8in]{\includegraphics[width=0.045\linewidth]{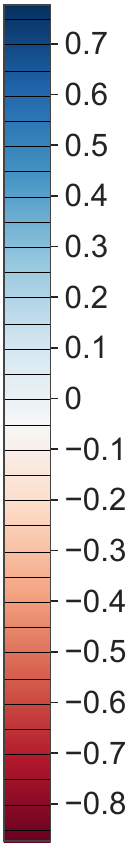}}
\end{tabular} \label{fig:joint-effect-cont-action-optimized}}

\caption{\footnotesize\textit{Contour plots representing the joint component functions $\hat{f}_{j}(\mathbf{s}_j,a)$ of $Q^\pi(\mathbf{s},a)$ estimated under the behavioral policy $\pi=\pi_b$ vs. the improved policy $\pi=\pi^*$. Each sub-figure is associated with the same nonparametric additive model, but represents a different state feature $\mathbf{s}_j$ on the y-axis.}}
\label{fig:joint-effect-cont-action}
\end{figure}

\textit{Joint effects between State Features and Actions.} In Figures \ref{fig:joint-effect-cont-action-surface} and \ref{fig:joint-effect-cont-action}, we examine the joint effects $\hat{f}_{j}(\mathbf{s}_j, a)$ between select state features $\mathbf{s}_j$ and the continuous action $a$ of the estimated action-value functions. When $a = \text{\textit{step count}}$ and $\mathbf{s} = \text{\textit{time spent at home}}$, we observed that increased time spent at home beyond 15 hours is associated with a positive increase in negative pain response across observed values of step count under the behavioral policy. This trend changes under the improved policy, where the joint effect is maximized when step count increases and time spent at home decreases and when step count is minimized and time spent at home increases. This suggests the joint effect of the improved policy may prioritize two distinct recovery strategies: patients can either maintain higher activity levels with less time at home or focus on rest with reduced activity.

Similar to the discrete action model, we observe a differential change in the joint effect associated with age. Under the improved and behavioral policy, an increase in step count across age is associated with an relative increase in the negative pain response. 

\section{Discussion}  
\label{Discussion}
In this study, we introduce a flexible, nonparametric additive representation of action-value functions. Our approach, KSH-LSPI, distinguishes itself from LSPI by avoiding parametric assumptions regarding the action-value function's form beyond additivity. It achieves this by incorporating ideas directly from local kernel regression and spline methods. This estimation approach affords KSH-LSPI the ability to capture the nonlinear additive contribution of each state action feature represented in the model. Furthermore, by introducing a group Lasso penalty to our primary objective function, we perform component-wise variable selection and retrieve a parsimonious representation of the action-value function. As demonstrated in our simulation, this approach can achieve competitive performance against modern neural network methods, particularly in scenarios with limited training data, while maintaining interpretability.

In the simulation study, we evaluated the performance of the proposed estimator and examined its sensitivity to changes in its hyperparameters. Future work aims to delve deeper, examining the estimator's finite sample properties both theoretically and through further simulations. The application of the proposed method to the spine disease dataset also provides new insights into mobilization behaviors that support postoperative pain management and reaffirmed several well-studied clinical findings. In future applications to spine disease recovery, we hope to extend the model by including categorical features such as gender, race, and diagnosis, as well as additional clinical features such as medication use. 

Our empirical analyses also revealed several methodological limitations. The method exhibits sensitivity to the discount factor, where higher values lead to estimation instability due to increased weighting of future states. The performance of the method depends on the selection of candidate features, where suboptimal choices can result in performance inferior to neural network approaches, an effect that becomes more pronounced as the training data increases. In regions of sparse data, component function values trend toward zero, an effect exacerbated by the group Lasso penalty and smaller bandwidth selections. Although the method effectively captures general functional shapes in well-sampled regions, it exhibits bias when estimating more complex underlying functions. Additionally, we found notable dependencies on the selected basis function, where trigonometric basis functions display greater sensitivity to state dimensions compared to B-spline bases and require stronger regularization for higher discount factors.

Furthermore, our application is not without limitations. The results of our model require careful interpretation and should not be deemed significant without comprehensive uncertainty quantification. In future adaptations of the KSH-LSPI model, we hope to formalize our uncertainty concerning our model estimates by incorporating a form of interval estimation. In the offline reinforcement learning setting, uncertainty-based approaches have shown promise in offline RL by prioritizing risk adverse policies when performing policy improvement \citep{Sonabend-W2020Expert-SupervisedEvaluation, ODonoghue2017TheExploration, Ghavamzadeh2016BayesianSurvey}. This naturally brings light to a limitation concerning our method's approach for policy improvement. After evaluating the current policy using KSH-LSTDQ, our policy improvement step greedily selects actions that maximize the estimated action-value function. Unfortunately, function approximation methods in offline reinforcement learning are prone to providing overly optimistic values for state action pairs that are unobserved in the training data. Hence, safe policy improvement steps, within the actor-critic framework, that regularize the learned policy toward the behavioral policy is encouraged in offline reinforcement learning, especially in healthcare applications \citep{Wang2020CriticRegression}. Another potential remedy would be to initialize our algorithm using an initial policy that closely reflects behaviors that would be suggested by a clinical expert. Initialization using physician-guided policies helps prevent the algorithm from becoming overly optimistic by selecting best actions that physicians themselves may select \citep{Gottesman2019GuidelinesHealthcare}.   

The push for interpretability in machine learning models, especially within healthcare contexts, is driven by a need for transparency in decision-making processes. Compared to the powerful, but often less interpretable neural network methodologies, nonparametric additive models for value-functions offers a representation where decision-making policies can be understood and scrutinized. Such interpretability is essential for potential clinical applications, given the need for clinicians to trust and validate the recommendations derived from these decision-making models. In conclusion, the KSH-LSPI model, although it has areas that require further refinement, provides a promising framework that aligns with the demand for both efficacy and transparency. 

\begin{acks}[Acknowledgments]
The authors would like to thank the anonymous referees, the Associate
Editor and the Editor for their constructive feedback and input.
\end{acks}

\begin{funding}
Junwei Lu is supported by NSF DMS 2434664, William F. Milton Fund, NIH/NCI R35CA220523, and NIH/NINDS R01NS098023.
\end{funding}

\begin{supplement}
\stitle{Supplement A -- Appendices}
\sdescription{We provide appendices containing additional information regarding the simulation study and the hyperparameter selection for the application to surgical recovery.}
\end{supplement}
\begin{supplement}
\stitle{Supplement B -- Python Code}
\sdescription{We provide a python implementation of the KSH-LSPI algorithm to reproduce the simulations and analysis performed in the paper. This code is provided as a supplement and is also available at \url{https://github.com/patricknnamdi/ksh-lspi}.}
\end{supplement}

\bibliographystyle{imsart-nameyear} 
\bibliography{references}

\end{document}